\documentclass[3p,onecolumn,authoryear,review]{elsarticle}
\usepackage{graphicx}
\usepackage{amsmath,amssymb,amsfonts}
\usepackage{amsthm}  
\usepackage{dsfont}
\usepackage{tikz}
\usepackage{pgfplots}\pgfplotsset{compat=1.16}\newlength{\figurewidth}\newlength{\figureheight}
\usepgfplotslibrary{groupplots}
\usepackage[nolist,nohyperlinks]{acronym}
\usepackage{tabularx}
\usepackage{booktabs}\newcommand{\otoprule}{\midrule[\heavyrulewidth]}
\usepackage[capitalize]{cleveref}
\crefname{figure}{Figure}{Figures}
\usepackage[per-mode=symbol]{siunitx}
\theoremstyle{researchquestion}\newtheorem{researchquestion}{Research question}
\Crefname{researchquestion}{Research question}{Research questions}

\usepackage{url}
\usepackage{natbib}
\usepackage{letltxmacro}
\LetLtxMacro{\autocite}{\citep}
\LetLtxMacro{\textcite}{\citet}

\makeatletter
\newcommand*{\org@overidelabel}{}
\let\org@overridelabel\@verridelabel
\renewcommand*{\@verridelabel}[1]{%
  \@bsphack
  \protected@write\@auxout{}{\string\AC@undonewlabel{#1@cref}}%
  \org@overridelabel{#1}%
  \@esphack
}%
\makeatother

\title{A Quantitative Method to Determine What Collisions Are Reasonably Foreseeable and Preventable}
\author[1]{Erwin de Gelder\corref{cor1}}
\ead{erwin.degelder@tno.nl}
\author[1]{Olaf Op den Camp}
\address[1]{TNO, Integrated Vehicle Safety, Helmond, The Netherlands}

\cortext[cor1]{Corresponding author}
\date{}

\let\originalleft\left
\let\originalright\right
\renewcommand{\left}{\mathopen{}\mathclose\bgroup\originalleft}
\renewcommand{\right}{\aftergroup\egroup\originalright}
\newcommand{\accelerationsymbol}{a}

  \newcommand{\accelerationaverage}{\bar{\accelerationsymbol}}
\newcommand{\bandwidth}{h}
  \newcommand{\bandwidthis}{\bandwidth_{\mathrm{NIS}}}

\renewcommand{\binom}[3]{\mathrm{Bin}\left(#1,#2,#3\right)}
  \newcommand{\binomk}{k}
  \newcommand{\binomn}{n}
  \newcommand{\binomp}{p}
\newcommand{\cardinality}[1]{\left|#1\right|}
\newcommand{\collisionprob}[1]{C\left(#1\right)}
  \newcommand{\collisionthreshold}{C_{\mathrm{p}}}

\newcommand{\densitysymbol}{f}
  
  \newcommand{\densitysub}[2]{\densitysymbol_{#2}\left(#1\right)}
  \newcommand{\densityissymbol}{g}
  \newcommand{\densityis}[1]{\densityissymbol\left(#1\right)}
  
\newcommand{\dimension}{d}
\newcommand{\dummyvar}{u}
  \newcommand{\dummyvarb}{v}
\newcommand{\e}[1]{\exp\left\{ #1 \right\}}
\newcommand{\entry}[2]{\left(#1\right)_{#2}}
\newcommand{\erf}[1]{\mathrm{erf}\left\{#1\right\}}
\newcommand{\excess}{y}
  \newcommand{\excessindex}{l}
  \newcommand{\excessinstance}[1]{\excess_{#1}}
  \newcommand{\excessindicesset}[1]{L\left(#1\right)}
\newcommand{\expectation}[1]{\mathds{E}\left[ #1 \right]}

\newcommand{\fromdistribution}{\sim}

\newcommand{\gapsymbol}{g}

  \newcommand{\gapinit}{\gapsymbol_{0}}
\newcommand{\gpd}[1]{G\left(#1\right)}
  \newcommand{\gpdscale}{\beta}
  \newcommand{\gpdshape}{\gamma}
  \newcommand{\gpddummy}{z}

\newcommand{\identitymatrix}[1]{I_{#1}}
\newcommand{\indexsampling}{j}
  \newcommand{\indexentry}{j}

\newcommand{\kdefunc}[1]{\hat{\densitysymbol}_{#1}}
  \newcommand{\kde}[2]{\kdefunc{#1}\left(#2\right)}
  \newcommand{\kdeint}[2]{\hat{F}_{#1}\left(#2\right)}
\newcommand{\kernelfunc}[1]{K \left( #1 \right)}

\newcommand{\meanis}{\mu_{\mathrm{NIS}}}
  
  \newcommand{\meanmc}{\mu_{\mathrm{MC}}}
\newcommand{\normtwo}[1]{\left\Vert #1 \right\Vert_2}
\newcommand{\numberofencounters}{n}
  \newcommand{\numberofencounterssc}[1]{\numberofencounters_{#1}}
  \newcommand{\numberofmc}{N_{\mathrm{MC}}}
  \newcommand{\numberofis}{N_{\mathrm{NIS}}}
  \newcommand{\numberofcritical}{N_{\mathrm{C}}}
  \newcommand{\numberofsimulations}{N_{\mathrm{sim}}}
  \newcommand{\numberofcollisions}[1]{N_{\mathrm{col}}\left(#1\right)}

\newcommand{\parameters}{x}
  
  \newcommand{\parametersinstance}[1]{\parameters_{#1}}
  \newcommand{\parameterordered}[1]{\parameters_{\left(#1\right)}}
  \newcommand{\parameterslower}{\parameters_{\mathrm{L}}}
  \newcommand{\parametersupper}{\parameters_{\mathrm{U}}}

\newcommand{\probability}[1]{\mathds{P}\left( #1 \right)}
  
\newcommand{\realnumbers}{\mathds{R}}

  \newcommand{\scenarioindex}{i}
  \newcommand{\scenariosnumberof}{N}
\newcommand{\scenariocategory}{\mathcal{C}}

  \newcommand{\scenariocategorylvd}{\scenariocategory_{\mathrm{LVD}}}
  \newcommand{\scenariocategoryci}{\scenariocategory_{\mathrm{cut-in}}}
  \newcommand{\scenariocategoryasv}{\scenariocategory_{\mathrm{ASV}}}

\newcommand{\simulationoutcome}[1]{R\left(#1\right)}
\newcommand{\speedsymbol}{v}

  \newcommand{\speedleadsymbol}{\speedsymbol_{\mathrm{l}}}

  \newcommand{\speeddifference}{\Delta_{\speedsymbol}}
  \newcommand{\speedinitego}{\speedsymbol_{\mathrm{e},0}}
  \newcommand{\speedinitlead}{\speedsymbol_{\mathrm{l},0}}
  
\newcommand{\stdis}{\sigma_{\mathrm{NIS}}}
  
  \newcommand{\stdmc}{\sigma_{\mathrm{MC}}}

\newcommand{\threshold}{u}
  \newcommand{\thresholdforeseeable}{\epsilon_{\mathrm{F}}}
  \newcommand{\thresholdpreventablecertainty}{\alpha}

\newcommand{\transpose}{^{\mkern-1.5mu\mathsf{T}}}
\newcommand{\ud}{\mathrm{\,d}}

\begin{acronym}[AAAAAAAA]
	\acro{alks}[ALKS]{Automated Lane Keeping System}\acroindefinite{alks}{an}{an}
	\acro{asv}[ASV]{Approaching Slower Vehicle}\acroindefinite{asv}{an}{an}
	\acro{av}[AV]{Automated Vehicle}\acroindefinite{av}{an}{an}
    \acro{ads}[ADS]{Automated Driving System}\acroindefinite{ads}{an}{an}
    \acro{cdf}[CDF]{Cumulative Distribution Function}
    \acro{evt}[EVT]{Extreme Value Theory}
    \acro{gpd}[GPD]{Generalized Pareto Distribution}
    \acro{idmplus}[IDM+]{Intelligent Driver Model plus}\acroindefinite{idmplus}{an}{an}
    \acro{kde}[KDE]{Kernel Density Estimation}
    \acro{lvd}[LVD]{Leading Vehicle Decelerating}\acroindefinite{lvd}{an}{a}
    \acro{nieon}[NIEON]{Non-Impaired Eyes ON the conflict}\acroindefinite{nieon}{an}{a}
	\acro{odd}[ODD]{Operational Design Domain}\acroindefinite{odd}{an}{an}
	\acro{pdf}[PDF]{Probability Density Function}
	\acro{ttc}[TTC]{Time to Collision}
\end{acronym}

\newcommand{\cstart}{}
\newcommand{\cend}{}

\begin{document}

\newpage
\begin{abstract}
	The development of \acp{ads} has made significant progress in the last years. 
	To enable the deployment of \acp{av} equipped with such \acp{ads}, regulations concerning the approval of these systems need to be established. 
	In 2021, the World Forum for Harmonization of Vehicle Regulations has approved a new United Nations regulation concerning the approval of \acp{alks}. 
	An important aspect of this regulation is that ``the activated system shall not cause any collisions that are reasonably foreseeable and preventable.'' 
	The phrasing of ``reasonably foreseeable and preventable'' might be subjected to different interpretations and, therefore, this might result in disagreements among \ac{av} developers and the authorities that are requested to approve \acp{av}. 
	
	The objective of this work is to propose a method for quantifying what is ``reasonably foreseeable and preventable''. 
	The proposed method considers the \ac{odd} of the system and can be applied to any \ac{odd}. 
	Having a quantitative method for determining what is reasonably foreseeable and preventable provides developers, authorities, and the users of \acp{ads} a \cstart better \cend understanding of the residual risks to be expected when deploying these systems in real traffic. 
	
	Using our proposed method, we can \cstart estimate \cend what collisions are reasonably foreseeable and preventable. 
	This will help in setting requirements regarding the safety of \acp{ads} and can lead to stronger justification for design decisions and test coverage for developing \acp{ads}. 
	
	\vspace{1em}
\end{abstract}
\maketitle
\acresetall

\section{Introduction}
\label{sec:introduction}

It is generally expected that \acp{ads} will make traffic safer by eliminating human errors, enable more comfortable rides, and reduce traffic congestion \autocite{chan2017advancements}.
Lower levels of automation systems, such as adaptive cruise control \autocite{mahdinia2020safety} and lane keeping assist systems \autocite{mammeri2015design}, are already widely deployed in modern cars and trucks.
Since the development of \acp{ads} has made significant progress, it is expected that \acp{ads} addressing higher levels of automation and covering the full dynamic driving task, i.e., SAE level 3 or higher \autocite{sae2021j3016}, are soon to be introduced on public roads \autocite{bimbraw2015autonomous, milakis2016scenarios, madni2018autonomous}.

The main idea of \iac{ads} is to either assist the driver with its driving task or to take over certain driving tasks. 
For higher levels of automation, the \ac{ads} may (temporarily) take over the full driving task of the driver.
A potential hindrance to the deployment of higher levels of automation is the Vienna Convention on road traffic of 1968. 
This convention, which is the basis of the national traffic laws of a large number of countries, requires that a human driver is in charge of driving \autocite{vellinga2019automated}.
Given the time at which the Vienna Convention has been drafted, it is not unreasonable that it has been assumed that a human is always in charge of the driving task.
Clearly, new regulations concerning the approval of higher levels of automation are needed.

To address the lack of a notion of an automated system that takes over the driving task of a human driver in the Vienna Convention, in 2021, the World Forum for Harmonization of Vehicle Regulations has approved a new United Nations regulation concerning the approval of an \ac{ads}, with the title ``uniform provisions concerning the approval of vehicles with regard to automated lane keeping systems'' \autocite{ece2021WP29}.
Regarding the system safety and fail-safe response, the following requirement is mentioned \autocite[Chapter~5]{ece2021WP29}:
``The activated system shall not cause any collisions that are reasonably foreseeable and preventable.''
This requirement leaves room for different interpretations, because the terms ``reasonably foreseeable'' and ``preventable'' have not been quantified. 
The different interpretations might result in disagreements among \ac{ads} developers and the authorities that are requested to approve \iac{ads}.

To provide more clarity on the terms ``reasonably foreseeable'' and ``preventable'', several studies are published. 
\textcite{schoener2020challenging} provides examples of scenarios that are (not) foreseeable and in which collisions are (not) preventable. 
This gives some directions, although the terms are not further quantified.
In the regulation itself, it is mentioned that ``the threshold for preventable/unpreventable is based on the simulated performance of a skilled and attentive human driver'' \autocite[Chapter~3]{ece2021WP29}.
This is further exploited by \textcite{mattas2022driver}, where four different driver behavior models are considered in parameterized scenarios and in each scenario, it is checked whether the driver could avoid a collision.
Two of these four driver behavior models are also presented in an amendment of the aforementioned regulation, see \autocite[Annex 3, Clauses 3.3 and 3.4]{ece2022WP29}.
\cstart In \autocite{kusano2022collision}, the collision avoidance performance of the Waymo \ac{ads} is compared with that of \iac{nieon} reference behavior model. \cend
\textcite{nakamura2022defining, muslim2023cut} provide a method to determine the range of the parameter values of ``foreseeable'' scenarios. 
This is done by fitting parametric distributions to the scenario parameter values and determining a range of the values that capture, e.g., \SI{99}{\percent}, of the probability mass.
\cstart In \autocite{kusano2022collision}, based on the analysis of driving data from instrumented Waymo vehicles and a set of so-called core scenarios identified earlier, new scenarios are created. 
These new scenarios may not have been observed in the driving data, but are still deemed to be reasonably foreseeable such that they are considered during the safety evaluation. \cend

We assume that ``reasonably foreseeable'' refers to the scenarios that are potentially leading to a collision. 
If a scenario is reasonably foreseeable, it is expected from the developers to consider this scenario during the design process and, therefore, ensure that the \ac{ads} safely handles such a scenario so that no collision occurs in this scenario.
Hence, this work addresses the following research question:
\begin{researchquestion}
	\label{rq:foreseeable}
	How to determine what are reasonably foreseeable scenarios?
\end{researchquestion}

To answer this research question, we follow a similar approach as \autocite{nakamura2022defining, muslim2023cut}, i.e., based on scenarios extracted from real-world data, we estimate the range of the scenario parameters that are reasonably foreseeable. To estimate this range, \iac{pdf} of the scenario parameters is estimated.
This range of parameters is such that the probability of encountering a scenario with parameter values outside this range is below a threshold.
Two different approaches are used for estimating the \ac{pdf} of the scenario parameters.
The first approach employs the non-parametric \ac{pdf} estimation technique called \ac{kde}.
Therefore, in contrast to the work of \textcite{nakamura2022defining, muslim2023cut}, no assumptions are needed regarding a parametric distribution and the independence of the scenario parameters.
The second approach uses only part of the data that is above a certain threshold, such that the \ac{gpd} can be assumed according to the \ac{evt}.
Both approaches come with advantages and disadvantages which will be discussed in this work.

The other term in the aforementioned requirement of the regulation \autocite{ece2021WP29} that leaves room for different interpretations is ``reasonably preventable''.
To make this more measurable, this work also proposes a method to answer the following question:
\begin{researchquestion}
	\label{rq:preventable}
	How to determine to which extent collisions are reasonably preventable?
\end{researchquestion}

We propose two alternative approaches that follow the \ac{alks} regulation that mentions that ``the threshold for preventable/unpreventable is based on the simulated performance of a skilled and attentive human driver'' \autocite[Chapter~3]{ece2021WP29}.
The first approach is to study to which extent a skilled and attentive human avoids a collision considering a certain \emph{type of scenario}.
This is estimated using Monte Carlo simulations while sampling the scenario parameters from the \ac{pdf} estimated using \ac{kde}.
Importance sampling is used to reduce the number of required simulations.
The second approach is to study to which extent a skilled and attentive human avoids a collision considering a certain \emph{scenario with specific parameter values}.
This approach leads to a range of parameter values for which the probability that a skilled and attentive human avoids a collision is above a certain threshold.
Of the two proposed approaches, the latter is mostly in line with the method proposed by \textcite{mattas2022driver} and in the annexes of \autocite{ece2022WP29}, but our approach differs in that we consider different outcomes when a single scenario is simulated multiple times due to the stochastic nature of the simulations. 
Whereas in \autocite{mattas2022driver}, a scenario either results in a collision or not, in our case, each scenario has a probability of a collision between 0 and 1.
We also provide a method to assess the certainty of the result, which can be used as a stop criterion for conducting further simulations.

This work is organized as follows. 
\Cref{sec:method} proposes the two methods to answer \cref{rq:foreseeable,rq:preventable}.
To illustrate the proposed methods and as a proof of concept, a case study is presented in \cref{sec:results}.
After the discussion in \cref{sec:discussion}, this work is concluded in \cref{sec:conclusions}.

\section{Methodology}
\label{sec:method}

\Cref{tab:definitions} presents the definitions of the terms that are used in our proposed method.
In this work, the probability of $\dummyvar$ is denoted by $\probability{\dummyvar}$, while $\expectation{\dummyvar}$ denotes the expectation of $\dummyvar$.
In the following subsection, we propose a method for answering \cref{rq:foreseeable}.
Next, \cref{sec:method preventable} presents a method for answering \cref{rq:preventable}.

\begin{table}
	\caption{Terms and definitions that are used in this work.}
	\label{tab:definitions}
	\begin{tabularx}{\linewidth}{lX}
		\toprule
		Term & Definition \\ \otoprule
		
		ODD & Operating conditions under which a given driving automation system or feature thereof is specifically designed to function, including, but not limited to, environmental, geographical, and time-of-day restrictions, and/or the requisite presence or absence of certain traffic or roadway characteristics \autocite{sae2021j3016} \\
		
		Scenario & Quantitative description of the relevant characteristics and activities and/or goals of the ego vehicle(s), the static environment, the dynamic environment, and all events that are relevant to the ego vehicle(s) within the time interval between the first and the last relevant event \autocite{degelder2020ontology} \\
		
		Scenario category & Qualitative description of the relevant characteristics and activities and/or goals of the ego vehicle(s), the static environment, and the dynamic environment \autocite{degelder2020ontology} \\
		
		Reasonably foreseeable & Likelihood of encountering in real-life is above a certain threshold \\
		
		Preventable & Avoidable by a skilled and attentive human driver \\
		\bottomrule
	\end{tabularx}
\end{table}

\subsection{Determining what reasonably foreseeable means}
\label{sec:method foreseeable}

To answer \cref{rq:foreseeable}, this section proposes a novel method for quantifying the set of scenarios that are reasonably foreseeable. 
To quantify ``reasonably foreseeable'', we use a threshold denoted by $\thresholdforeseeable$.
This threshold $\thresholdforeseeable$ has units ``per hour''.
We propose a method to look for a parameter range such that the likelihood of encountering scenarios with its parameters outside this parameter range equals $\thresholdforeseeable$.
This proposed method consists of three steps:
\begin{enumerate}
	\item Identify the scenarios that are part of the so-called \ac{odd} of the \ac{ads}.
	\item Determine the exposure of these scenarios, i.e., the expected number of occurrences per hour of driving.
	\item Determine the probability of encountering scenarios within a specified parameter range.
\end{enumerate}
The first two steps are explained in \cref{sec:scenario identification,sec:exposure}, respectively.
Two different approaches for the third step are presented in \cref{sec:probability parameter range,sec:extreme value theory}.

\subsubsection{Identification of scenarios}
\label{sec:scenario identification}

An \ac{ads} is designed to operate within its \ac{odd}, which is defined by the \ac{ads} developer and typically consists of a geofence and some known operational conditions, e.g., see \autocite{gm2018selfdriving, aurora2019newera, waymo2021safety}.
Once deployed, the \ac{ads} needs to deal with many scenarios and the \ac{odd} in which the \ac{ads} is operating determines the variety of these scenarios.
To determine the reasonably foreseeable scenarios, the \ac{odd} needs to be known.

Considering the wide variety of scenarios, we propose to distinguish between quantitative scenarios and qualitative scenarios, where scenario categories refer to the latter, see \cref{tab:definitions}.
It is assumed that all possible scenarios within a given \ac{odd} can be categorized into one or more scenario categories. 
This assumption does not limit the applicability of the methodology proposed in this work, though it might require many scenario categories to describe all these scenarios.
In the remainder of this section, we propose a method to determine the set of scenarios within a scenario category $\scenariocategory$, where $\scenariocategory$ denotes a scenario category.
For example, a scenario category could refer to all cut-in scenarios in the \ac{odd} of the \ac{ads}.
See \autocite{degelder2019scenariocategories} for more examples of scenario categories.

\subsubsection{Probability of exposure}
\label{sec:exposure}

To determine the exposure, we estimate the expected number of encounters of a scenario that belongs to scenario category $\scenariocategory$.
Let $\numberofencounterssc{\scenariocategory}$ denote the number of encounters per unit of time of a scenario belonging to scenario category $\scenariocategory$. 
We express the exposure as $\expectation{\numberofencounterssc{\scenariocategory}}$, i.e., the expected number of encounters per unit of time of a scenario belonging to scenario category $\scenariocategory$: 
\begin{equation}
	\label{eq:exposure}
	\expectation{\numberofencounterssc{\scenariocategory}}
	 = \sum_{\numberofencounters=1}^{\infty} \numberofencounters \probability{\numberofencounterssc{\scenariocategory}=\numberofencounters},
\end{equation}
where $\probability{\numberofencounterssc{\scenariocategory}=\numberofencounters}$ denotes the probability of encountering $\numberofencounters$ scenarios belonging to scenario category $\scenariocategory$ in one hour of driving. 

We propose to determine $\probability{\numberofencounterssc{\scenariocategory}=\numberofencounters}$, $\numberofencounters=0,1,2,\ldots$, based on data, because the data provide a quantitative way to estimate $\probability{\numberofencounterssc{\scenariocategory}=\numberofencounters}$.
Furthermore, assuming  that the data are collected with the same conditions as specified by the \ac{odd} of the \ac{ads}, the data provides an objective way to estimate $\probability{\numberofencounterssc{\scenariocategory}=\numberofencounters}$. 
The probability can be estimated by counting the number of occurrences of the scenarios in the data.
The method to find the scenarios belonging to $\scenariocategory$ is explained in \autocite{degelder2020scenariomining}:
First, tags are used to describe activities, such as lane changing and braking, and statuses, such as ``leading vehicle'' and ``driving slower''.
Note that a tag is typically associated with an object and has a start time and an end time. 
Second, by searching for a particular combination of these tags that describes the scenario category $\scenariocategory$, the start and end time of the scenarios are found.

\subsubsection{Estimate parameter range with estimated probability density function}
\label{sec:probability parameter range}

Let $\parameters\in\realnumbers^{\dimension}$  denote the $\dimension$-dimensional parameter vector that describes a scenario. 
To estimate the probability that $\parameters$ is within a the range $[\parameterslower, \parametersupper]$, we need to know the \ac{cdf} of $\parameters$. 
To estimate the \ac{cdf}, we will first estimate the \ac{pdf}.
Let $\densitysub{\cdot}{\scenariocategory}$ denote the \ac{pdf} of the parameters of the scenarios from scenario category $\scenariocategory$. 
Typically, the \ac{pdf} $\densitysub{\cdot}{\scenariocategory}$ is unknown, so it needs to be estimated.
To estimate the \ac{pdf}, we use the observed scenarios that have also been used to estimate the exposure (\cref{sec:exposure}).
Since the shape of the \ac{pdf} is also unknown beforehand, assuming a predefined functional form of the \ac{pdf} for which certain parameters are fitted to the data may lead to inaccurate fits unless a lot of hand-tuning is applied. 
As an alternative, we propose to estimate $\densitysub{\cdot}{\scenariocategory}$ using \ac{kde} \autocite{rosenblatt1956remarks, parzen1962estimation}.
Let $\left\{\parametersinstance{\scenarioindex}\right\}_{\scenarioindex=1}^{\scenariosnumberof}$ denote the set of parameters of the $\scenariosnumberof$ observed scenarios that belong to the scenario category $\scenariocategory$, then the density $\densitysub{\cdot}{\scenariocategory}$ is estimated by
\begin{equation}
	\label{eq:kde}
	\kde{\scenariocategory}{\parameters}
	= \frac{1}{\scenariosnumberof\bandwidth^{\dimension}}\sum_{\scenarioindex=1}^{\scenariosnumberof}
	\kernelfunc{\frac{1}{\bandwidth}\left(\parameters-\parametersinstance{\scenarioindex}\right)}.
\end{equation}
Here, $\kernelfunc{\cdot}$ is the so-called kernel and $\bandwidth$ is the bandwidth.
The choice of the kernel function is not as important as the choice of the bandwidth \autocite{turlach1993bandwidthselection, duong2007ks}.
We use the often-used Gaussian kernel, which is given by
\begin{equation}
	\label{eq:gaussian kernel}
	\kernelfunc{\dummyvar} = \frac{1}{\left(2\pi\right)^{\dimension/2}} 
	\e{-\frac{1}{2} \normtwo{\dummyvar}^2},
\end{equation}
where $\normtwo{\dummyvar}^2=\dummyvar\transpose\dummyvar$ denotes the squared 2-norm of $\dummyvar$.

The bandwidth $\bandwidth>0$ is a free parameter that controls the width of the kernel.
A larger bandwidth results in a smoother \ac{pdf}, but choosing $\bandwidth$ too large may result in loss of details in the \ac{pdf}.
Methods for estimating the bandwidth range from simple reference rules like Silverman's rule of thumb \autocite{silverman1986density} to more elaborate methods \autocite{turlach1993bandwidthselection}.
We use leave-one-out cross validation to determine the optimal bandwidth because this minimizes the Kullback-Leibler divergence between the estimated \ac{pdf}, $\kde{\scenariocategory}{\cdot}$, and the unknown \ac{pdf} that underlies the data, $\densitysub{\cdot}{\scenariocategory}$ \autocite{turlach1993bandwidthselection}.
Because the same amount of smoothing is applied in every direction, the data are first normalized such that the standard deviation equals 1 for each of the $\dimension$ parameters.

Using the estimated \ac{pdf} of the form of \cref{eq:kde}, the estimated \ac{cdf} is obtained by integrating $\kde{\scenariocategory}{\parameters}$. 
More explicitly, the estimated \ac{cdf} equals:
\begin{equation}
	\kdeint{\scenariocategory}{\parameters}
	= \int_{-\infty}^{\entry{\parameters}{1}} \cdots \int_{-\infty}^{\entry{\parameters}{\dimension}}
	\kde{\scenariocategory}{\parameters}
	\ud\entry{\parameters}{1} \cdots \ud\entry{\parameters}{\dimension},
\end{equation}
where $\entry{\parameters}{\indexentry}$ denotes the $\indexentry$-th entry of the vector $\parameters$.
Using the Gaussian kernel of \cref{eq:gaussian kernel}, this gives
\begin{equation}
	\kdeint{\scenariocategory}{\parameters} = 
	\frac{1}{\scenariosnumberof} \sum_{\scenarioindex=1}^{\scenariosnumberof}
	\prod_{\indexentry=1}^{\dimension}
	\left[ \frac{1}{2} + \frac{1}{2} \erf{\frac{\entry{\parameters}{\indexentry} - \entry{\parametersinstance{\scenarioindex}}{\indexentry}}{\bandwidth\sqrt{2}}} \right],
\end{equation}
where $\erf{\cdot}$ is the error function that is defined as
\begin{equation}
	\erf{\dummyvar} = \frac{2}{\sqrt{\pi}} \int_{0}^{\dummyvar} \e{-\dummyvarb^2} \ud \dummyvarb.
\end{equation}
The probability $\probability{\parameters\in[\parameterslower, \parametersupper]}$, i.e., $\parameters$ is within the range $[\parameterslower, \parametersupper]$, is estimated using the evaluation of the estimated \ac{cdf}, $\kdeint{\scenariocategory}{\cdot}$, at the vertices of the hyperrectangle spanned by $\parameterslower$ and $\parametersupper$.
In case of $\dimension=1$, this would simply be:
\begin{equation}
	\probability{\parameters\in[\parameterslower, \parametersupper]}
	\approx \kdeint{\scenariocategory}{\parametersupper} - \kdeint{\scenariocategory}{\parameterslower}.
\end{equation}
By solving the following equation with respect to $\parameterslower$ and $\parametersupper$, the range of parameters that are reasonably foreseeable can be determined:
\begin{equation}
	\label{eq:foreseeable}
	\expectation{\numberofencounterssc{\scenariocategory}} \cdot \left(1 - \probability{\parameters\in[\parameterslower, \parametersupper]} \right) = \thresholdforeseeable.
\end{equation}
Here, $\thresholdforeseeable$ is a threshold with units ``per hour'', such that the probability of encountering a scenario of scenario category $\scenariocategory$ with its parameters outside the range $[\parameterslower, \parametersupper]$ equals $\thresholdforeseeable$. 
Note that typically multiple solutions for $\parameterslower$ and $\parametersupper$ exist.

\subsubsection{Estimate parameter range using extreme values}
\label{sec:extreme value theory}

It might be difficult to justify a particular distribution of the scenario parameters.
Therefore, as explained in \cref{sec:probability parameter range}, a non-parametric method for estimating the \ac{pdf} of the scenario parameters is adopted, such that there is no need to assume a particular shape of the \ac{pdf}.
Nevertheless, our second approach for determining the parameter range makes use of a parametric distribution, namely the \acf{gpd}.
As we will explain next, according to the \acf{evt}, this choice is justified as long as we only use the data that is above a certain threshold \autocite{franke2004statistics}.

\begin{figure}
	\centering
	\input{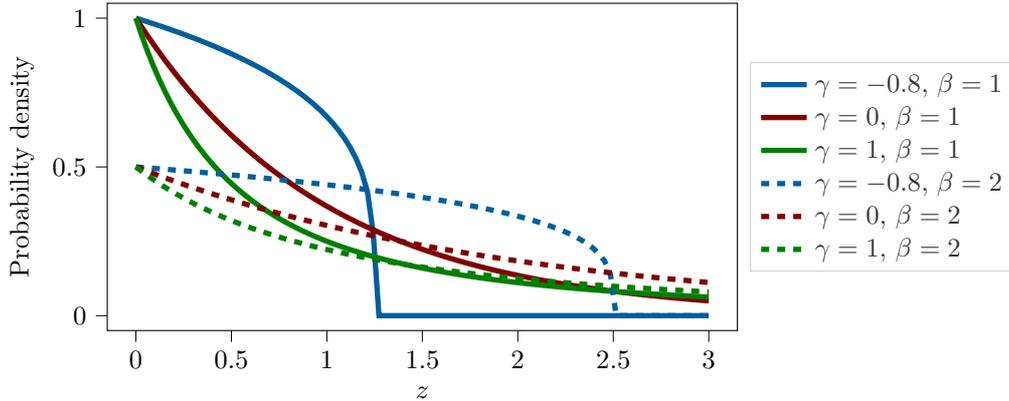}
	\caption{Probability density of the \ac{gpd}, i.e., the derivative of \cref{eq:gpd}, for different values of the shape parameter, $\gpdshape$, and the scale parameters $\gpdscale$.}
	\label{fig:gpd}
\end{figure}

Just as in \cref{sec:probability parameter range}, let us assume that we have a set of parameter values, denoted by $\left\{\parametersinstance{\scenarioindex}\right\}_{\scenarioindex=1}^{\scenariosnumberof}$, with $\parametersinstance{\scenarioindex}\in\realnumbers^{\dimension}$.
For now, let us assume that $\dimension=1$.
We will discuss the generalization toward multiple parameters at the end of this section.
Let $\threshold$ denote the threshold such that we only consider the parameters that are larger than $\threshold$.
To describe this mathematically, let
\begin{equation}
	\excessindicesset{\threshold} = \left\{\scenarioindex ; \parametersinstance{\scenarioindex}>\threshold \right\}
\end{equation}
denote the set of indices for which the parameter values are larger than $\threshold$.
Based on this, we can define a set of ``excess values'':
\begin{equation}
	\left\{ \excessinstance{\excessindex} \right\}_{\excessindex=1}^{\cardinality{\excessindicesset{\threshold}}}
	= \left\{ \parametersinstance{\scenarioindex} - \threshold \right\}_{\scenarioindex \in \excessindicesset{\threshold}}.
\end{equation}
Here, $\cardinality{\excessindicesset{\threshold}}$ denotes the cardinality of $\excessindicesset{\threshold}$, i.e., the number of parameter values of the set $\left\{\parametersinstance{\scenarioindex}\right\}_{\scenarioindex=1}^{\scenariosnumberof}$ that are above the threshold $\threshold$.
Following \autocite{franke2004statistics}, under a mild assumption\footnote{The distribution of $\parameters$ must be contained in the maximum domain of attraction of the generalized extreme value distribution. For more details, see \autocite{balkema1974residual, pickands1975statistical}.} regarding the underlying distribution of $\parameters$, the distribution of the excess value $\excess$ approaches the \ac{gpd} as $\threshold\rightarrow\infty$.
The \ac{gpd} is given by:
\begin{equation}
	\label{eq:gpd}
	\gpd{\gpddummy} =
	\begin{cases}
		1 - \left(1 + \frac{\gpdshape \gpddummy}{\gpdscale} \right)^{\frac{-1}{\gpdshape}} & \text{if } \gpdshape \neq 0 \\
		1 - \e{-\frac{\gpddummy}{\gpdscale}} & \text{if } \gpdshape=0
	\end{cases},
\end{equation}
where the support is $\gpddummy \geq 0$ for $\gpdshape \geq 0$ and $0 \leq \gpddummy \leq -\gpdscale/\gpdshape$ for $\gpdshape < 0$.
The two parameters of the \ac{gpd} are the so-called shape parameter, $\gpdshape$, and the scale parameter, $\gpdscale>0$.
\Cref{fig:gpd} shows the probability density of the \ac{gpd} for different values of $\gpdshape$ and $\gpdscale$.

\setlength{\figurewidth}{.6\linewidth}
\setlength{\figureheight}{.6\figurewidth}

To determine the parameter range using the \ac{gpd}, we simply assume that there is no lower bound or $\parameterslower=-\infty$.
The upper bound is found using the following equation:
\begin{equation}
	\label{eq:foreseeableb}
	\expectation{\numberofencounterssc{\scenariocategory}} \cdot
	\probability{\parameters > \threshold} \cdot
	\left(1 - \gpd{\parametersupper - \threshold} \right) 
	= \thresholdforeseeable.
\end{equation}
Here, the probability $\probability{\parameters > \threshold}$ can simply be estimated by dividing the total number of parameter values that exceed $\threshold$ by the total number of scenarios in scenario category $\scenariocategory$, i.e.,
\begin{equation}
	\probability{\parameters > \threshold}
	= \frac{ {\cardinality{\excessindicesset{\threshold}}}}{\scenariosnumberof}.
\end{equation}
To use \cref{eq:foreseeableb}, $\threshold$ must be chosen such that $\expectation{\numberofencounterssc{\scenariocategory}} \cdot \probability{\parameters > \threshold} \geq \thresholdforeseeable$. 
In practice, this is not a problem because $\thresholdforeseeable$ is generally small.
Nevertheless, if it would result in a too low value of $\threshold$, one can simply estimate the upper bound using $\expectation{\numberofencounterssc{\scenariocategory}} \cdot \probability{\parameters > \parametersupper} = \thresholdforeseeable$.

The reason why we have assumed that $\dimension=1$ is that, in that case, it is straightforward to order the data.
For example, it is straightforward to determine which scenario parameters are larger than the threshold $\threshold$.
When considering multiple parameters per scenario ($\dimension>1$), one option is to only estimate an upper bound for one parameter and assume that all possible values for the remaining parameters are reasonably foreseeable.
A second option is to assume independence of the different parameters.
The approach presented in this section can easily be expanded to consider multiple independent parameters.
A third option is to use a function for mapping the $\dimension$-dimensional parameter space to a one-dimensional parameter space.
In that case, the presented approach can be used to find an upper bound in the one-dimensional parameter space and this upper bound can be mapped to the $\dimension$-dimensional parameter space using the inverse of the function that is used for the mapping.

\subsection{Determining what reasonably preventable means}
\label{sec:method preventable}

To answer \cref{rq:preventable}, this section proposes a method for quantifying the extent to which collisions are preventable. 
This is achieved by considering a skilled and attentive human driver as a benchmark for \iac{ads}. 
Therefore, we will estimate the probability that a skilled and attentive human driver is able to prevent collisions.
In \cref{sec:preventable scenario category}, we first estimate to which extent a skilled and attentive human driver can avoid collisions considering all scenarios within a scenario category.
We will employ Monte Carlo simulations using importance sampling to estimate this probability \autocite{degelder2021risk}.
As an alternative approach, in \cref{sec:preventable scenario}, we estimate to which extent a skilled and attentive human driver can avoid a collision considering a specific scenario.

\subsubsection{Collision probability considering a scenario category}
\label{sec:preventable scenario category}

As already introduced in \cref{sec:probability parameter range}, it is assumed that the scenarios of a scenario category $\scenariocategory$ can be described using a $\dimension$-dimensional parameter vector $\parameters\in\realnumbers^{\dimension}$.
Let the (stochastic) outcome of a simulation of a scenario with parameters $\parameters$ be denoted by by $\simulationoutcome{\parameters}$, where ${\simulationoutcome{\parameters}=1}$ means that the simulation of the scenario with parameters $\parameters$ ends with a collision and otherwise ${\simulationoutcome{\parameters}=0}$. 
To estimate the extent to which a skilled and attentive human driver can avoid collisions considering all scenarios of a scenario category, we estimate the expectation of $\simulationoutcome{\parameters}$, i.e., $\expectation{\simulationoutcome{\parameters}}$.
The skilled and attentive human driver is modeled with a driver behavior model, see \cref{sec:driver model} for more details.
To estimate this expectation, the \ac{pdf} of $\parameters$, denoted by $\kde{\scenariocategory}{\cdot}$, needs to be estimated.
For this, the \ac{kde} of \cref{eq:kde} is used.
With (crude) Monte Carlo sampling, $\expectation{\simulationoutcome{\parameters}}$ is estimated by sampling the parameters from the \ac{pdf} and averaging the results of the simulation runs.
Sampling from $\kde{\scenariocategory}{\cdot}$ is straightforward.
First, an integer $\indexsampling\in\{1,\ldots,\scenariosnumberof\}$ is chosen randomly with each integer having equal likelihood.
Next, a random sample is drawn from a Gaussian with covariance $\bandwidth^2\identitymatrix{\dimension}$ and mean $\parametersinstance{\indexsampling}$, where $\identitymatrix{\dimension}$ denotes the $\dimension$-by-$\dimension$ identity matrix.

With crude Monte Carlo, the probability of a collision is calculated by taking the mean of $\simulationoutcome{\parameters}$ over a large number, $\numberofmc$, of different realizations of $\parameters$:
\begin{equation}
	\label{eq:monte carlo}
	\expectation{\simulationoutcome{\parameters}} \approx
	\meanmc = \frac{1}{\numberofmc} \sum_{\indexsampling=1}^{\numberofmc}
	\simulationoutcome{\parametersinstance{\indexsampling}}, 
	\quad \parametersinstance{\indexsampling} \fromdistribution \kdefunc{\scenariocategory}.
\end{equation}

It is easy to see that the crude Monte Carlo result is unbiased, i.e., $\expectation{\meanmc}=\expectation{\simulationoutcome{\parameters}}$.
To estimate the potential approximation error, $\meanmc-\expectation{\simulationoutcome{\parameters}}$, the estimated standard deviation of \cref{eq:monte carlo} is commonly used:
\begin{equation}
	\label{eq:mc std}
	\stdmc = \frac{1}{\numberofmc} 
	\sqrt{\sum_{\indexsampling=1}^{\numberofmc} \left( \meanmc - \simulationoutcome{\parametersinstance{\indexsampling}} \right)^2}.
\end{equation}

In general, it can be expected that the probability of collision, $\expectation{\simulationoutcome{\parameters}}$, is small.
As a result, none or few of the $\numberofmc$ scenario simulations may end with a collision and the relative uncertainty, i.e., $\stdmc/\meanmc$, will be high or undefined (in case none of the scenario simulations ends with a collision).
With importance sampling, the scenario parameters are sampled from a different distribution --- the so-called importance density --- such that the simulation runs focus more on scenarios in which the probability of collision is high.
This will lead to a lower relative uncertainty of the estimated probability of collision.
We use nonparametric importance sampling, which means that a nonparametric method is used to estimate the unknown optimal importance density \autocite{zhang1996nonparametric}. 
More specifically, as proposed in \autocite{zhang1996nonparametric, deGelder2017assessment}, the nonparametric importance sampling method employs \ac{kde} to construct the importance density.

In particular, this work uses the importance density $\densityis{\cdot}$, which is defined as follows:
\begin{equation}
	\label{eq:importance density}
	\densityis{\parameters} = \frac{1}{\numberofcritical \bandwidthis^{\dimension}}
	\sum_{\scenarioindex=1}^{\numberofcritical}
	\kernelfunc{\frac{1}{\bandwidthis}\left(\parameters-\parameterordered{\scenarioindex}\right)}.
\end{equation}
Here, $\{\parameterordered{1},\ldots,\parameterordered{\numberofcritical}\}$ denote the $\numberofcritical$ ``most critical'' scenarios of the crude Monte Carlo sampling ($\numberofcritical<\numberofmc$).
We use the $\numberofcritical$ scenarios that resulted in the lowest minimum \ac{ttc} \autocite{hayward1972near} during each simulation run to determine $\numberofcritical$ ``most critical'' scenarios. 
The \ac{ttc} is defined as the ratio of the distance toward an approaching object and the speed difference with that object.
The bandwidth $\bandwidthis$ is estimated using leave-one-out cross validation.
If $\numberofis$ scenarios are simulated in nonparametric importance sampling, the estimated probability of collision is
\begin{equation}
	\label{eq:is}
	\expectation{\simulationoutcome{\parameters}} \approx
	\meanis = \frac{1}{\numberofis} \sum_{\indexsampling=1}^{\numberofis}
	\simulationoutcome{\parametersinstance{\indexsampling}}
	\frac{\kde{\scenariocategory}{\parametersinstance{\indexsampling}}}{\densityis{\parametersinstance{\indexsampling}}},
	\quad \parametersinstance{\indexsampling} \fromdistribution \densityissymbol.
\end{equation}
The weight $\kde{\scenariocategory}{\parametersinstance{\indexsampling}}/\densityis{\parametersinstance{\indexsampling}}$ is used to correct for the bias introduced by sampling from $\densityis{\cdot}$ instead of $\kde{\scenariocategory}{\cdot}$.
Because $\densityis{\parameters}>0$ whenever $\simulationoutcome{\parameters}\kde{\scenariocategory}{\parameters}>0$, \cref{eq:is} gives unbiased results \autocite{owen2013montecarlo}.
The estimated standard deviation of \cref{eq:is} is
\begin{equation}
	\label{eq:std is}
	\stdis = \frac{1}{\numberofis}
	\sqrt{\sum_{\indexsampling=1}^{\numberofis} \left( \frac{\simulationoutcome{\parametersinstance{\indexsampling}}\kde{\scenariocategory}{\parametersinstance{\indexsampling}}}{\densityis{\parametersinstance{\indexsampling}}} - \meanis \right)^2}.
\end{equation}
For more details regarding the nonparametric importance sampling, the reader is referred to \autocite{degelder2021risk}.

\subsubsection{Collision probability considering a scenario}
\label{sec:preventable scenario}

To determine the probability of a collision considering a specific scenario with parameters $\parameters$, we can simulate that specific scenario.
As introduced in \cref{sec:preventable scenario category}, the outcome of a simulation of a scenario with parameters $\parameters$ is denoted by $\simulationoutcome{\parameters}$, where $\simulationoutcome{\parameters}=1$ indicates that the simulation ended with a collision and $\simulationoutcome{\parameters}=0$ otherwise.
Let $\collisionprob{\parameters}$ denote the probability that a skilled and attentive human driver does not prevent a collision.
If the simulation is fully deterministic, a single simulation suffices.
If this is not the case, a straightforward way to compute the collision probability is to repeat a certain number of simulations with the same $\parameters$ and count the number of simulations that result in a collision. 
If $\numberofsimulations$ denotes the number of simulations, then the probability of a collision can be estimated using
\begin{equation}
	\label{eq:preventable scenario}
	\collisionprob{\parameters} = \frac{\numberofcollisions{\parameters}}{\numberofsimulations},
\end{equation}
where $\numberofcollisions{\parameters}=\sum_{\indexsampling=1}^{\numberofsimulations} \simulationoutcome{\parameters}$ denotes the number of simulation runs that end in a collision.

To determine whether it is reasonable to assume that a human cannot avoid a collision, we are interested in the cases where $\collisionprob{\parameters}>\collisionthreshold$, where $\collisionthreshold$ denotes a certain threshold.
Therefore, the exact value of $\collisionprob{\parameters}$ is less relevant.
This knowledge is used to limit the number of simulations that needs to be conducted for each scenario with parameters $\parameters$.
The number of simulations, $\numberofsimulations$, is increased until the certainty that $\collisionprob{\parameters}$ is below or above $\collisionthreshold$ is more than $1-\thresholdpreventablecertainty$, where $\thresholdpreventablecertainty$ denotes the probability of a wrong conclusion regarding $\collisionprob{\parameters}>\collisionthreshold$.
This certainty is computed using the binomial distribution.
In other words, $\numberofsimulations$ is increased until one of the following conditions is reached:
\begin{align}
	\sum_{\binomk=0}^{\numberofcollisions{\parameters}} \binom{\binomk}{\collisionthreshold}{\numberofsimulations} & < \thresholdpreventablecertainty, \label{eq:nsim threshold a} \\
	\sum_{\binomk=\numberofcollisions{\parameters}}^{\numberofsimulations} \binom{\binomk}{\collisionthreshold}{\numberofsimulations} & < \thresholdpreventablecertainty, \label{eq:nsim threshold b}
\end{align}
with
\begin{equation}
	\binom{\binomk}{\binomp}{\binomn} = \frac{\binomn!}{\binomk!\,\left(\binomn-\binomk\right)!} \binomp^{\binomk} \left(1-\binomp\right)^{\binomn-\binomk}.
\end{equation}
In practice, using \cref{eq:nsim threshold a,eq:nsim threshold b} with $\collisionthreshold=0.5$ and $\thresholdpreventablecertainty=0.01$, it means that a minimum of seven simulation runs are conducted.
To limit the number of simulation runs, especially when $\collisionprob{\parameters}\approx\collisionthreshold$, $\numberofsimulations$ is further limited to 100.

\section{Analysis and Results}
\label{sec:results}

To illustrate the proposed method for determining in a quantitative manner what are reasonably foreseeable scenarios and preventable collisions, the method is applied in a case study.
This case study considers three scenario categories that are detailed in \cref{sec:example scenarios}.
Details regarding the used data set are mentioned in \cref{sec:data set}.
In this case study, a model is used to describe the driver behavior, see \cref{sec:driver model}.
Finally, \cref{sec:results foreseeable,sec:results preventable} present the results regarding the reasonably foreseeable scenarios and the preventable collisions, respectively.

\subsection{Scenarios and parameterization}
\label{sec:example scenarios}

This case study considers three scenario categories.
The first scenario category is named ``\ac{lvd}'' and denoted by $\scenariocategorylvd$.
This is one of the two scenario categories that are mentioned as possibly ``critical scenarios'' in the \ac{alks} regulation \autocite{ece2021WP29}.
The scenarios that belong to this scenario category consider a leading vehicle that is driving in front of the ego vehicle.
Initially, this leading vehicle is driving at a constant speed $\speedinitlead>0$.
The leading vehicle decelerates such that its end speed is $\speedinitlead-\speeddifference$, with $0<\speeddifference<\speedinitlead$.
The average deceleration of the leading vehicle is denoted by $\accelerationaverage>0$.
To model the speed of the leading vehicle during its deceleration activity, a sinusoidal shape is assumed.
It is further assumed that the ego vehicle is initially driving behind the leading vehicle at a constant distance and speed and that the desired speed of the ego vehicle equals the initial speed.
The parameter vector that describes \ac{lvd} scenarios contains the initial speed of the leading vehicle ($\speedinitlead$), the ratio of the speed difference and the initial speed ($\speeddifference/\speedinitlead$), and the average deceleration ($\accelerationaverage$):
\begin{equation}
	\label{eq:lvd parameters}
	\parameters\transpose =
	\begin{bmatrix}
		\speedinitlead & \speeddifference/\speedinitlead & \accelerationaverage
	\end{bmatrix}.
\end{equation}

Note that because of the infinite support of the Gaussian kernel of \cref{eq:gaussian kernel}, the kernel density estimator will assign a positive density for parameter values that are invalid. 
For the first parameter, this is insignificant, so the \ac{pdf} is simply set to zero for $\speedinitlead\leq0$.
The resulting \ac{pdf} is rescaled such that it still integrates to $1$.
For the other two parameters, however, we have mapped the parameter before applying the \ac{kde}, such that there is no positive density for parameter values that are invalid.
First, instead of using $\speeddifference/\speedinitlead$, which is between $0$ and $1$, we use $-\log\left(\speedinitlead/\speeddifference-1\right)$ and for $\accelerationaverage$, which is larger than $0$, $\log \accelerationaverage$ is used.
In the remainder of this work, although we use the mapped parameters for the density estimation, the results and figures show the parameters of \cref{eq:lvd parameters} as these are easier to interpret than the mapped parameters.

The second scenario category is named ``cut-in'' and denoted by $\scenariocategoryci$. 
This scenario category is the other scenario category that is mentioned as a potentially ``critical scenario'' in the \ac{alks} regulation \autocite{ece2021WP29}.
Cut-in scenarios consider a vehicle that suddenly cuts into the lane of the ego vehicle such that this vehicle becomes the leading vehicle of the ego vehicle.
At the moment of the cut-in, the longitudinal gap between the two vehicles is $\gapinit>0$.
For similar reasons as explained for the \ac{lvd} scenario, for the \ac{kde}, $\log \gapinit$ is used instead of $\gapinit$.
It is assumed that the vehicle that performs the cut-in drives at a constant speed which is denoted by $\speedleadsymbol>0$.
The initial speed of the ego vehicle is denoted by $\speedinitego>0$.
The parameter vector that describes cut-in scenarios contains initial gap ($\gapinit$), the initial speed of the ego vehicle ($\speedinitego$), and the ratio of the initial speeds of the leading vehicle and the ego vehicle ($\speedleadsymbol/\speedinitego$):
\begin{equation}
	\parameters\transpose =
	\begin{bmatrix}
		\gapinit & \speedinitego & \speedleadsymbol/\speedinitego
	\end{bmatrix}.
\end{equation}

The third scenario category is named ``\ac{asv}'' and denoted by $\scenariocategoryasv$.
This scenario category \ac{asv}, which also includes scenarios in which the leading vehicle is stationary at standstill, accounts for more than \SI{25}{\percent} of all crashes that involve two vehicles in the U.S.A.\ \autocite{USDoT2007precrashscenarios}.
In \iac{asv} scenario, the ego vehicle is initially driving at a speed $\speedinitego>0$ while approaching a vehicle that is moving at constant speed of $0 \leq \speedleadsymbol < \speedinitego$.
The parameter vector that describes \ac{asv} scenarios contains the initial speed of the ego vehicle ($\speedinitego$) and the ratio of the speeds of the leading vehicle and the ego vehicle ($\speedleadsymbol/\speedinitego$):
\begin{equation}
	\parameters\transpose =
	\begin{bmatrix}
		\speedinitego & \speedleadsymbol/\speedinitego
	\end{bmatrix}.
\end{equation}

\subsection{Data set}
\label{sec:data set}

To estimate the exposure and the \acp{pdf} of the scenario parameters, the data set described in \autocite{paardekooper2019dataset6000km} is used.
The data were recorded from a single vehicle in which 20 experienced drivers were asked to drive a prescribed route. 
Each driver drove the \SI{50}{km} route six times, which resulted in \SI{63}{\hour} of data.
The route contains urban roads, rural roads, and highways.
To measure the surrounding traffic, the vehicle was equipped with three radars and one camera. 
The surrounding traffic was measured by fusing the data of the radars and the camera as described in \autocite{elfring2016effective}.

To extract the scenarios from the data set, the approach described in \autocite{degelder2020scenariomining} is used.
\ac{lvd} scenarios are found by querying for a vehicle that has the tags ``leading vehicle'' and ``decelerating'' at the same time.
Cut-in scenarios are found by querying for a vehicle that initially has the tags ``changing lane'' and ``no leading vehicle'' which changes into the tags ``changing lane'' and ``leading vehicle'' \autocite{degelder2020scenariomining}.
\ac{asv} scenarios are found by querying for a vehicle that has the tags ``leading vehicle'' and ``driving slower''.
In 63 hours of driving, 1300 \ac{lvd} scenarios, 297 cut-in scenario, and 291 \ac{asv} scenarios have been found.

\subsection{Modeling of the driver behavior}
\label{sec:driver model}

To model the driver behavior, the \ac{idmplus} \autocite{schakel2010effects} is used. 
The parameters of the \ac{idmplus} model are adopted from \autocite{treiber2000congested} except for the desired time headway, which is set to \SI{1.2}{\second} \autocite{schakel2010effects}.
The \ac{idmplus} is a so-called collision-free model, because collisions are always prevented, sometimes at the costs of a high deceleration and zero reaction time.
To prevent the use of unrealistic reaction times, we assume that the driver has a reaction time that is distributed according to a log-normal distribution with a mean of \SI{0.92}{\second} and a standard deviation of \SI{0.28}{\second} \autocite{green2000long}.
In addition, the maximum braking capacity is set to \SI{6}{\meter\per\second\squared}.
Furthermore, it is assumed that the driver does not respond to vehicles that are further than \SI{150}{\meter} away.

\subsection{Results for reasonably foreseeable scenarios}
\label{sec:results foreseeable}

\Cref{fig:foreseeable pdf lvd} shows the results for the \ac{lvd} scenarios using the approach outlined in \cref{sec:probability parameter range}.
The data, which are shown by the histograms, are used to estimate the \ac{pdf} using \ac{kde}.
Because 1300 \ac{lvd} scenarios have been observed in 63 hours of driving, it follows that $\expectation{\numberofencounterssc{\scenariocategorylvd}} =\SI[per-mode=reciprocal]{20.6}{\per\hour}$.
Therefore, if $\thresholdforeseeable=\SI[per-mode=reciprocal]{0.1}{\per\hour}$, it follows from \cref{eq:foreseeable} that we look for the range $[\parameterslower, \parametersupper]$ such that $\probability{\parameters\in[\parameterslower, \parametersupper]}=0.9951$.
The red vertical lines in \cref{fig:foreseeable pdf lvd} show one possibility for $\parameterslower$ and $\parametersupper$ such that $\probability{\parameters\in[\parameterslower, \parametersupper]}=0.9951$.
Here, $\parameterslower\transpose=\begin{bmatrix}\SI{0}{\meter\per\second} & 0 & \SI{0}{\meter\per\second\squared}\end{bmatrix}$ and $\parametersupper\transpose=\begin{bmatrix}\SI{45.0}{\meter\per\second} & 1 & \SI{3.01}{\meter\per\second\squared}\end{bmatrix}$.
With $\thresholdforeseeable=\SI[per-mode=reciprocal]{0.01}{\per\hour}$, the same lower bound could be used with an upper bound $\parametersupper\transpose=\begin{bmatrix}\SI{50.0}{\meter\per\second} & 1 & \SI{5.41}{\meter\per\second\squared}\end{bmatrix}$, such that $\probability{\parameters\in[\parameterslower, \parametersupper]}=0.9995$.

\newlength{\histwidth}\setlength{\histwidth}{.45\linewidth}
\newlength{\histheight}\setlength{\histheight}{.6\histwidth}
\setlength{\figurewidth}{\histwidth}
\setlength{\figureheight}{\histheight}
\begin{figure}
	\centering
	\input{figs/pdf_lvd.tikz}
	\caption{Result for reasonably foreseeable \ac{lvd} scenarios using the approach of \cref{sec:probability parameter range}.
		The original data are represented by the histograms.
		The blue lines denote the estimated marginal distributions based on the \ac{kde} of \cref{eq:kde}.
		The red vertical lines show the boundaries of the reasonably foreseeable parameter range with $\thresholdforeseeable=\SI[per-mode=reciprocal]{0.1}{\per\hour}$ and the green vertical lines show these boundaries with $\thresholdforeseeable=\SI[per-mode=reciprocal]{0.01}{\per\hour}$ (if different).}
	\label{fig:foreseeable pdf lvd}
\end{figure}

\Cref{fig:foreseeable gpd lvd} shows the result for \ac{lvd} scenarios using the approach outlined in \cref{sec:extreme value theory}.
We only look at the upper bound of the average deceleration.
The threshold $\threshold=\SI{1.18}{\meter\per\second\squared}$ is chosen such that \SI{10}{\percent} of the data is used for the set of excess values, i.e., $\probability{\accelerationaverage > \threshold}=0.1$.
\Cref{fig:foreseeable gpd lvd} shows the histogram of these excess values.
Fitting the parameters of the \ac{gpd} of \cref{eq:gpd} on these excess values using the maximum likelihood method resulted in $\gpdshape=0.051$ and $\gpdscale=\SI{0.36}{\meter\per\second\squared}$.
Using \cref{eq:foreseeableb} with $\thresholdforeseeable=\SI[per-mode=reciprocal]{0.1}{\per\hour}$ results in an upper bound of $\accelerationaverage=\SI{2.34}{\meter\per\second\squared}$, see the red vertical line in \cref{fig:foreseeable gpd lvd}.
In other words, based on this approach with the threshold $\thresholdforeseeable=\SI[per-mode=reciprocal]{0.1}{\per\hour}$, all \ac{lvd} scenarios with $\accelerationaverage \leq \SI{2.34}{\meter\per\second\squared}$ are reasonably foreseeable and all \ac{lvd} scenarios with $\accelerationaverage > \SI{2.34}{\meter\per\second\squared}$ are not reasonably foreseeable.
With $\thresholdforeseeable=\SI[per-mode=reciprocal]{0.01}{\per\hour}$, the upper bound shifts to $\accelerationaverage=\SI{3.36}{\meter\per\second\squared}$.
Note that using this approach, the other scenario parameters, $\speedinitlead$ and $\speeddifference/\speedinitlead$, are not used to determine whether a scenario is reasonably foreseeable.

\newlength{\evtwidth}\setlength{\evtwidth}{.45\linewidth}
\newlength{\evtheight}\setlength{\evtheight}{.6\evtwidth}
\setlength{\figurewidth}{\evtwidth}
\setlength{\figureheight}{\evtheight}
\begin{figure}
	\centering
	\input{figs/gpd_lvd.tikz}
	\caption{Result for reasonably foreseeable \ac{lvd} scenarios using the approach of \cref{sec:extreme value theory}.
		The histogram represents \SI{10}{\percent} of the data with the highest average deceleration.
		The blue line is the probability density of the fitted \ac{gpd} of \cref{eq:gpd}.
		The red (green) vertical line at $\accelerationaverage=\SI{2.34}{\meter\per\second\squared}$ ($\accelerationaverage=\SI{3.36}{\meter\per\second\squared}$) denotes the upper bound if $\thresholdforeseeable=\SI[per-mode=reciprocal]{0.1}{\per\hour}$ ($\thresholdforeseeable=\SI[per-mode=reciprocal]{0.01}{\per\hour}$) is used.}
	\label{fig:foreseeable gpd lvd}
\end{figure}

\begin{figure}
	\centering
	\input{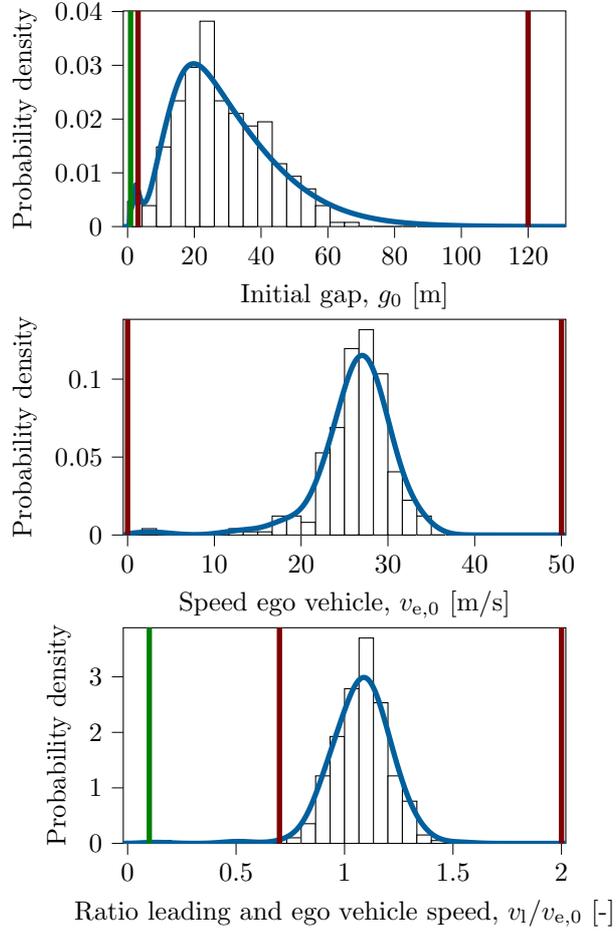}
	\caption{Result for reasonably foreseeable cut-in scenarios using the approach of \cref{sec:probability parameter range}.
		For a more detailed explanation, see \cref{fig:foreseeable pdf lvd}.}
	\label{fig:foreseeable pdf cutin}
\end{figure}

\Cref{fig:foreseeable pdf cutin} shows the results for the cut-in scenarios using the approach outlined in \cref{sec:probability parameter range}. 
In total, 297 cut-in scenarios have been observed, so $\expectation{\numberofencounterssc{\scenariocategoryci}} = 297 / \SI{63}{\hour} = \SI[per-mode=reciprocal]{4.71}{\per\hour}$.
So, with a threshold of $\thresholdforeseeable=\SI[per-mode=reciprocal]{0.1}{\per\hour}$, we look for the range $[\parameterslower, \parametersupper]$ such that $\probability{\parameters\in[\parameterslower, \parametersupper]}=0.9788$.
This probability is obtained using $\parameterslower\transpose=\begin{bmatrix}\SI{3.14}{\meter} & \SI{0}{\meter\per\second} & 0.700 \end{bmatrix}$ and $\parametersupper\transpose=\begin{bmatrix}\SI{120}{\meter} & \SI{50.0}{\meter\per\second} & 2.00 \end{bmatrix}$, see the red vertical lines in \cref{fig:foreseeable pdf cutin}.
When using a threshold $\thresholdforeseeable=\SI[per-mode=reciprocal]{0.01}{\per\hour}$, the parameter range of reasonably foreseeable scenarios becomes substantially larger with $\parameterslower\transpose=\begin{bmatrix}\SI{0.900}{\meter} & \SI{0}{\meter\per\second} &  0.100\end{bmatrix}$ and $\parametersupper\transpose=\begin{bmatrix}\SI{120}{\meter} & \SI{50.0}{\meter\per\second} & 2.00 \end{bmatrix}$.

\setlength{\figurewidth}{\histwidth}
\setlength{\figureheight}{\histheight}

\Cref{fig:foreseeable gpd cutin} shows the result for cut-in scenarios using the approach based on the \ac{evt}.
We look at the ratio of the leading vehicle speed and the initial ego vehicle speed, $\speedleadsymbol/\speedinitego$.
If this ratio is below one, the ego vehicle needs to perform an evasive action, e.g., by braking or steering.
Typically, for lower values of $\speedleadsymbol/\speedinitego$, a stronger evasive maneuver is expected.
Therefore, we are interested in the lower bound of $\speedleadsymbol/\speedinitego$ rather than the upper bound.
To use the approach as outlined in \cref{sec:extreme value theory}, we simply estimate the upper bound of $-\speedleadsymbol/\speedinitego$.
We use a threshold of $\threshold=-0.91$ because \SI{10}{\percent} of the values of $\speedleadsymbol/\speedinitego$ are below \num{0.91}.
\Cref{fig:foreseeable gpd cutin} shows the histogram of these excess values.
Note that the data and the upper bound of $-\speedleadsymbol/\speedinitego$ are (again) multiplied with $-1$, such that the x-axis represents $\speedleadsymbol/\speedinitego$ and the upper bound represents the lower bound of $\speedleadsymbol/\speedinitego$.
Fitting the parameters of the \ac{gpd} of \cref{eq:gpd} on these excess values using the maximum likelihood method resulted in $\gpdshape=0.62$ and $\gpdscale=\num{0.045}$.
Because we assume that $\speedleadsymbol/\speedinitego > 0$, the fitted \ac{pdf} is set to zero if $\speedleadsymbol/\speedinitego \leq 0$ and the \ac{pdf} is rescaled such that the \ac{pdf} integrates to 1.
Using \cref{eq:foreseeableb} with $\thresholdforeseeable=\SI[per-mode=reciprocal]{0.1}{\per\hour}$ results in a lower bound of $\speedleadsymbol/\speedinitego=\num{0.81}$, see the red vertical line in \cref{fig:foreseeable gpd lvd}.
With $\thresholdforeseeable=\SI[per-mode=reciprocal]{0.01}{\per\hour}$, the lower bound shifts to $\speedleadsymbol/\speedinitego=\num{0.42}$.

\setlength{\figurewidth}{\evtwidth}
\setlength{\figureheight}{\evtheight}
\begin{figure}
	\centering
	\input{figs/gpd_cutin.tikz}
	\caption{Result for reasonably foreseeable cut-in scenarios using the approach of \cref{sec:extreme value theory}.
		The histogram represents \SI{10}{\percent} of the data with the smallest ratio of leading vehicle speed and the initial ego vehicle speed, $\speedleadsymbol/\speedinitego$.
		The blue line is the probability density of the fitted \ac{gpd} of \cref{eq:gpd}.
		The red (green) vertical line at $\speedleadsymbol/\speedinitego=\num{0.81}$ ($\speedleadsymbol/\speedinitego=\num{0.42}$) denotes the lower bound if $\thresholdforeseeable=\SI[per-mode=reciprocal]{0.1}{\per\hour}$ ($\thresholdforeseeable=\SI[per-mode=reciprocal]{0.01}{\per\hour}$) is used.}
	\label{fig:foreseeable gpd cutin}
\end{figure}

\begin{figure}
	\centering
	\input{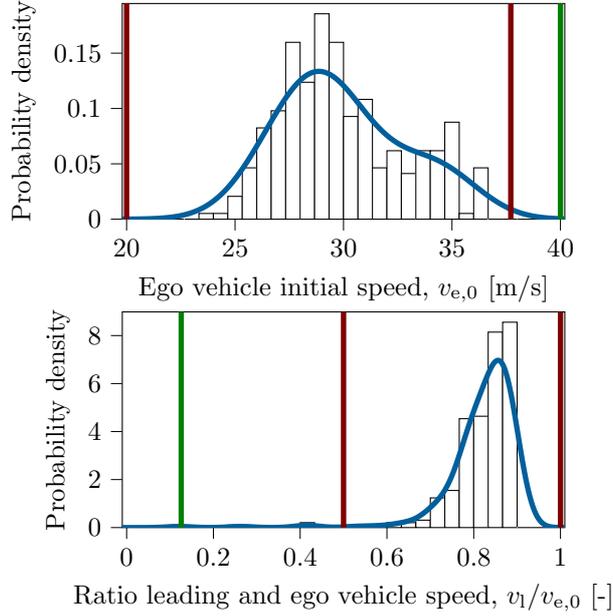}
	\caption{Result for reasonably foreseeable \ac{asv} scenarios using the approach of \cref{sec:probability parameter range}.
		For a more detailed explanation, see \cref{fig:foreseeable pdf lvd}.}
	\label{fig:foreseeable pdf asv}
\end{figure}

\Cref{fig:foreseeable pdf asv} shows the results for the \ac{asv} scenarios using the approach outlined in \cref{sec:probability parameter range}. 
In total, 291 \ac{asv} scenarios have been observed, so $\expectation{\numberofencounterssc{\scenariocategoryasv}} = 291 / \SI{63}{\hour} = \SI[per-mode=reciprocal]{4.63}{\per\hour}$.
One option for the range $[\parameterslower, \parametersupper]$ to satisfy \cref{eq:foreseeable} with $\thresholdforeseeable=\SI[per-mode=reciprocal]{0.1}{\per\hour}$ is $\parameterslower\transpose=\begin{bmatrix}\SI{20.0}{\meter\per\second} & 0.50\end{bmatrix}$ and $\parametersupper\transpose=\begin{bmatrix}\SI{37.7}{\meter\per\second} & 1.0\end{bmatrix}$, see the red vertical lines in \cref{fig:foreseeable pdf asv}.
With $\thresholdforeseeable=\SI[per-mode=reciprocal]{0.01}{\per\hour}$, the upper bound of the initial ego vehicle speed, $\speedinitego$, shifts to $\SI{40}{\meter\per\second}$ and the lower bound of the ratio of leading vehicle speed and the initial ego vehicle speed, $\speedleadsymbol/\speedinitego$, shifts to \num{0.13}, see the green vertical lines in \cref{fig:foreseeable pdf asv}.

\setlength{\figurewidth}{\histwidth}
\setlength{\figureheight}{\histheight}

For the approach using the \ac{evt}, the same steps as for the cut-in scenarios are followed, see \cref{fig:foreseeable gpd asv}.
This results in a lower bound of $\speedleadsymbol/\speedinitego=\num{0.62}$ with $\thresholdforeseeable=\SI[per-mode=reciprocal]{0.1}{\per\hour}$ and a lower bound of $\speedleadsymbol/\speedinitego=\num{0.27}$ with $\thresholdforeseeable=\SI[per-mode=reciprocal]{0.01}{\per\hour}$.

\setlength{\figurewidth}{\evtwidth}
\setlength{\figureheight}{\evtheight}
\begin{figure}
	\centering
	\input{figs/gpd_asv.tikz}
	\caption{Result for reasonably foreseeable \ac{asv} scenarios using the approach of \cref{sec:extreme value theory}.
		The histogram represents \SI{10}{\percent} of the data with the smallest ratio of leading vehicle speed and the initial ego vehicle speed, $\speedleadsymbol/\speedinitego$.
		The blue line is the probability density of the fitted \ac{gpd} of \cref{eq:gpd}.
		The red (green) vertical line at $\speedleadsymbol/\speedinitego=\num{0.62}$ ($\speedleadsymbol/\speedinitego=\num{0.27}$) denotes the lower bound if $\thresholdforeseeable=\SI[per-mode=reciprocal]{0.1}{\per\hour}$ ($\thresholdforeseeable=\SI[per-mode=reciprocal]{0.01}{\per\hour}$) is used.}
	\label{fig:foreseeable gpd asv}
\end{figure}

\subsection{Results for preventable collisions}
\label{sec:results preventable}

First, the probability that a human driver cannot avoid a collision in scenarios that belong to a specific scenario category is estimated using the approach outlined in \cref{sec:preventable scenario category}.
As explained in \cref{sec:preventable scenario category}, Monte Carlo simulations have been conducted.
Using the estimated \ac{pdf} of the scenario parameters (see \cref{fig:foreseeable pdf lvd,fig:foreseeable pdf cutin,fig:foreseeable pdf asv}) for sampling, $\numberofmc=10000$ simulations have been conducted.
The results of these simulations are reported in \cref{tab:preventable}.
From the 10000 \ac{lvd} scenarios, 9 scenarios resulted in a collision and 24 of the 10000 simulations of the cut-in scenarios has resulted in a collision.
None of the 10000 \ac{asv} scenarios resulted in a collision.
This illustrates the need for a more efficient sampling strategy, as also explained in \cref{sec:preventable scenario category}.
When applying importance sampling, the estimated probability of a collision in \iac{lvd} scenario is $7.81\cdot 10^{-4}$ with an estimated uncertainty of $5.51\cdot 10^{-5}$.
Similarly, the estimated probabilities of a collision in cut-in and \ac{asv} scenarios are $2.06\cdot 10^{-3}$ and $2.56\cdot 10^{-6}$, respectively.

\begin{table}
	\centering
	\caption{Probability that a human driver cannot avoid a collision.}
	\label{tab:preventable}
	\begin{tabular}{lllll}
		\toprule
		Scenario category &
		$\meanmc$ of \cref{eq:monte carlo} &
		$\stdmc$ of \cref{eq:mc std} &
		$\meanis$ of \cref{eq:is} &
		$\stdis$ of \cref{eq:std is} \\ \otoprule
		
		$\scenariocategorylvd$ & $9.00\cdot 10^{-4}$ & $3.00\cdot 10^{-3}$ & $7.81\cdot 10^{-4}$ & $5.51\cdot 10^{-5}$ \\ 
		$\scenariocategoryci$  & $2.40\cdot 10^{-3}$ & $4.89\cdot 10^{-4}$ & $2.06\cdot 10^{-3}$ & $6.53\cdot 10^{-5}$ \\ 
		$\scenariocategoryasv$ & $0.0$                & $0.0$                & $2.56\cdot 10^{-6}$  & $2.05\cdot 10^{-6}$  \\ 
		\bottomrule
	\end{tabular}
\end{table}

\begin{figure}
	\centering
	\input{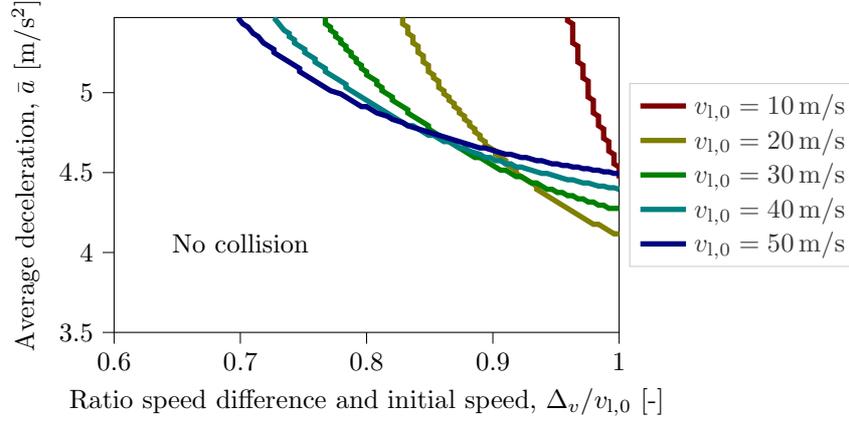}
	\caption{Lines that split the parameter space of the \ac{lvd} scenarios into a region where the estimated probability that a skilled and attentive human collides is smaller than $\collisionthreshold=0.5$ (i.e., $\collisionprob{\parameters}<\collisionthreshold$) (lower left region) and larger than $\collisionthreshold=0.5$ (i.e., $\collisionprob{\parameters}>\collisionthreshold$) (upper right region).}
	\label{fig:preventable lvd}
\end{figure}

\Cref{fig:preventable lvd} shows the result of reasonably preventable collisions in \ac{lvd} scenarios.
We have used the approach described in \cref{sec:preventable scenario} with $\collisionthreshold=0.5$ and $\thresholdpreventablecertainty=0.01$.
The parameters $\parameters$ for which the probability that a skilled and attentive human collides, i.e., $\collisionprob{\parameters}$, is estimated are taken from a grid with:
\begin{itemize}
	\item $100$ values of $\speeddifference/\speedinitlead$ ranging from zero to one;
	\item $100$ values of $\accelerationaverage$ ranging from \SI{3.5}{\meter\per\second\squared} to \SI{5.41}{\meter\per\second\squared}, where this latter value is the upper bound determined using the estimated \ac{pdf} with $\thresholdforeseeable=\SI[per-mode=reciprocal]{0.01}{\per\hour}$ (see \cref{fig:foreseeable pdf lvd}); and
	\item $\speedinitlead$ ranging from \SI{10}{\meter\per\second} to \SI{50}{\meter\per\second} in steps of \SI{10}{\meter\per\second}, where \SI{50}{\meter\per\second} is the upper bound determined using the estimated \ac{pdf} with $\thresholdforeseeable=\SI[per-mode=reciprocal]{0.01}{\per\hour}$ (see \cref{fig:foreseeable pdf lvd}).
\end{itemize}
Each line in \cref{fig:preventable lvd} shows where $\collisionprob{\parameters} \approx 0.5$ for a specific value of $\speedinitlead$.
In each case, the parameter region below the line is where the collision probability is below $0.5$ while the region above the line is where the collision probability is above $0.5$.
For example, if $\speeddifference/\speedinitlead=0.85$ and $\accelerationaverage=\SI{5}{\meter\per\second\squared}$, a collision is reasonably preventable in case $\speedinitlead=\SI{10}{\meter\per\second}$ or $\speedinitlead=\SI{20}{\meter\per\second}$ while a collision is not reasonably preventable with these values of $\speeddifference/\speedinitlead$ and $\accelerationaverage$ if $\speedinitlead=\SI{30}{\meter\per\second}$, $\speedinitlead=\SI{40}{\meter\per\second}$, or $\speedinitlead=\SI{50}{\meter\per\second}$.

\setlength{\figurewidth}{.5\linewidth}
\setlength{\figureheight}{.7\figurewidth}

\Cref{fig:preventable cutin} presents the result of reasonably preventable collision in cut-in scenarios, which are obtained in a similar way as the results of the \ac{lvd} scenarios in \cref{fig:preventable lvd}.
\Cref{fig:preventable cutin} shows that if $\speedinitego=\SI{50}{\meter\per\second}$, there is a large range for $\gapinit$ and $\speedleadsymbol/\speedinitego$ for which a collision is not reasonably preventable, i.e., the region on the left of the blue line.
For smaller values of $\speedinitego$, this region becomes smaller. 
For example, if $\speedinitego=\SI{10}{\meter\per\second}$, all cut-in scenarios with $\gapinit>\SI{20}{\meter}$ are reasonably preventable, i.e., the region above the red line in \cref{fig:preventable cutin}.

\begin{figure}
	\centering
	\input{figs/preventable_cutin.tikz}
	\caption{Lines that split the parameter space of the cut-in scenarios into a region where $\collisionprob{\parameters}<\collisionthreshold$ (upper right region) and $\collisionprob{\parameters}>\collisionthreshold$ (lower left region) with $\collisionthreshold=0.5$.}
	\label{fig:preventable cutin}
\end{figure}

We have also conducted simulations of the \ac{asv} scenarios.
Based on the parameter range determined using the estimated \ac{pdf} with $\thresholdforeseeable=\SI[per-mode=reciprocal]{0.01}{\per\hour}$ (see \cref{fig:foreseeable pdf asv}), simulations has been conducted with $\speedinitego$ ranging from \SI{20}{\meter\per\second} to \SI{40}{\meter\per\second} and $\speedleadsymbol/\speedinitego$ ranging from $0.13$ to one.
In all cases, the estimated probability that a skilled and attentive human driver does not prevent a collision is below $\collisionthreshold=0.5$.

\section{Discussion}
\label{sec:discussion}

This paper presented a method to determine which scenarios are reasonably foreseeable and to which extent collisions are preventable. 
\cstart Given the approved United Nations regulation that states that ``the activated system shall not cause any collisions that are reasonably foreseeable and preventable'' \autocite[Clause~5.1.1]{ece2021WP29}, the results of the presented method can be used to identify in which scenarios \iac{ads} is expected to avoid collisions. \cend
This section discusses the approaches and results that are presented and some directions for future research.
The discussion in \cref{sec:discussion foreseeable} is related to \cref{rq:foreseeable} and the discussion in \cref{sec:discussion preventable} is related to \cref{rq:preventable}.

\subsection{Discussion related to reasonably foreseeable scenarios}
\label{sec:discussion foreseeable}

Two different approaches to determine what are reasonably foreseeable scenarios have been presented.
Both approaches estimate a range for the parameters that describe the scenarios that belong to a specific scenario category.
The first approach (\cref{sec:probability parameter range}) estimates \iac{pdf} of the parameters using \ac{kde}.
An advantage of this approach is that no strong assumptions regarding the shape of the \ac{pdf} are considered. 
However, near the tails of the estimated \ac{pdf}, the influence of the choice of the kernel becomes more apparent.
The Gaussian kernel of \cref{eq:gaussian kernel} is opted because it makes evaluating the \ac{cdf} feasible.
Choosing a kernel from the family of multivariate beta kernels \autocite{duong2015spherically}, such as the multivariate uniform or Epanechnikov kernel, would require numerical integration.
This becomes very cumbersome, especially because a high numerical precision is required for low values of $\thresholdforeseeable$.
Apart from the practical advantage, there is no argument that justifies the choice of the Gaussian kernel.
Hence, if the estimated bounds are beyond the most extreme data points that are used to construct the kernel density estimator, then the precision of these bounds may be low.
In our case study, this is the case when $\thresholdforeseeable=\SI[per-mode=reciprocal]{0.01}{\per\hour}$ is used.

The alternative approach is based on the \ac{evt} (\cref{sec:extreme value theory}) and uses a fitted \ac{gpd} to determine the upper bound of a scenario parameter.
The advantage of this approach is that the \ac{evt} justifies the choice of the \ac{gpd}.
\Ac{evt} has successfully been used to determine extremes that are beyond the most extreme values in the data \autocite{dehaan1994extreme, tarko2012use}.
A disadvantage of this approach is that it is typically applied to only one parameter, although it is possible to extend it to multiple parameters, e.g., using a mapping of multiple parameters to one parameter. 

\cstart Similar as in \autocite{nakamura2022defining, muslim2023cut}, both approaches presented in this work construct \iac{pdf} which is used to estimate the range of the parameters of the reasonably foreseeable scenarios. 
Our proposed methods differ from the method used in \autocite{nakamura2022defining, muslim2023cut} because fewer assumptions regarding the \ac{pdf} estimation are made. 
In \autocite{nakamura2022defining, muslim2023cut}, it is assumed that the scenario parameters are independent, that the parameters are distributed according to the Beta distribution, and the lower and upper bounds of these Beta distributions are set to some assumed values. 
If we would use the method proposed in \autocite{nakamura2022defining, muslim2023cut}, we would have reported a different outcome. 
We cannot objectively state which outcome is better because there is no objective truth. 
In general, however, we can state that if the assumptions made in \autocite{nakamura2022defining, muslim2023cut} are correct, it is likely that the outcome based on their method would be more accurate. 
On the other hand, if there is no reason to make certain assumptions regarding the \ac{pdf}, it generally leads to more accurate results if one does not rely on those assumptions. 
For our case study, we cannot justify the assumptions made in \autocite{nakamura2022defining, muslim2023cut}.
Thus, for our case study, the methods proposed in this work might be more suitable. \cend

The proposed method for determining the reasonably foreseeable scenarios relies on data.
It is, therefore, important that the data are adequate.
This means that the data need to match the \ac{odd} of the \ac{ads}.
For example, the data that have been used in the case study were obtained from driving a specific route multiple times during daytime and good weather conditions.
If the \ac{odd} of the \ac{ads} considers the same route during daytime and good weather conditions, then the data are representing this \ac{odd}.
If, however, the \ac{odd} is substantially different, extra arguments are needed to justify that the data still represent the \ac{odd}.
The estimated exposure of the scenarios and the estimated parameter \acp{pdf} might not be accurate for the specified \ac{odd} in case the data have been recorded under different circumstances.
As a result, the estimated parameter range of the reasonably foreseeable scenarios might not be accurate enough.
The adequacy of the data also concerns the amount of data.
It is important that we have enough data to accurately determine the \ac{pdf} of the scenario parameters. 
As an order of magnitude, the number of hours of data must be roughly $1/\thresholdforeseeable$.

An important parameter for determining the parameter range of the reasonably foreseeable parameters is $\thresholdforeseeable$.
For the sake of illustration, in the case study, we have used $\thresholdforeseeable=\SI[per-mode=reciprocal]{0.1}{\per\hour}$ and $\thresholdforeseeable=\SI[per-mode=reciprocal]{0.01}{\per\hour}$, which means that, on average, one scenario in $10$ or $100$ hours of driving, respectively, is found with its parameters not within the range $[\parameterslower,\parametersupper]$.
When deploying \iac{ads} on a large scale, however, a much lower value of $\thresholdforeseeable$ might be better. 
This automatically results in a larger parameter range of the reasonably foreseeable scenarios.
The reason that the presented case study has used $\thresholdforeseeable=\SI[per-mode=reciprocal]{0.01}{\per\hour}$ as a minimum value is that the estimation of the reasonably foreseeable parameter range with $63$ hours of data would be inaccurate. 
Smaller values of $\thresholdforeseeable$ require more data to accurately determine the parameter range of the reasonably foreseeable scenarios.

\cstart Our work provides methods to determine the range of the parameter values of reasonably foreseeable scenarios for a given scenario category. 
As mentioned in \cref{sec:scenario identification}, it might require many scenario categories to cover a given \ac{odd}. 
It remains future work to define a method that identifies all reasonably foreseeable scenarios, i.e., across all relevant scenario categories.
The work of \textcite{kusano2022collision} already provides substantial work in this direction, as they have identified a large set of scenarios based on the combinations of so-called core scenarios and salient factors which are properties
of the actors or scenes that may impact the performance of the system. \cend

\subsection{Discussion related to reasonably preventable collisions}
\label{sec:discussion preventable}

In \cref{sec:method preventable}, two approaches are presented for studying the extent to which a collision is preventable.
The first approach considers all scenarios that belong to a specific scenario category.
A method is provided to calculate the probability that a skilled and attentive human driver cannot avoid a collision.
The result is a probability for a given scenario category that can be used by an \ac{av} developer as a benchmark. 
If \iac{av} reports a lower probability of colliding in scenarios for all relevant scenario categories, then the \ac{av} may be considered safer than a skilled and attentive human driver.
Note that it is still possible that the \ac{av} fails to avoid a collision in scenarios that a skilled and attentive human driver can handle safely: the collision probabilities consider all scenarios in a scenario category, so as long as the \ac{av} handles some scenarios in a scenario category better than a skilled and attentive human driver, the \ac{av} may handle other scenarios from the same scenario category worse without having a higher collision probability.
It remains an open question whether this is acceptable to authorities and the general public.

The second approach for studying the extent to which a collision is preventable considers individual scenarios.
Using this approach, it is expected that \iac{av} can safely deal with each scenario that a skilled and attentive human driver can safely handle too. This might be closer to the intuition that \iac{av} must prevent all collisions if it is reasonable to assume that a human driver can also prevent these collisions.
A potential disadvantage of this approach, compared to the first approach, is that it might take a long time to study, design, and validate \iac{av} that prevents all collisions that a skilled and attentive human driver can reasonably prevent.
This might delay the actual deployment of \acp{av} while the earlier introduction of \acp{av} --- that are potentially less safe --- might actually have a positive impact on the overall traffic safety.

\cstart Note that both approaches can complement each other and that a combination is also possible, as already proposed by \textcite{kusano2022collision}.
The acceptance criterion for \iac{ads} can be based on a comparison between the \ac{ads} and that (a model of) a skilled and attentive human driver --- similar to the second approach --- and, in addition, it can be required that the performance across all scenarios of a scenario category is at least comparable to a threshold --- similar to the first approach. \cend

The estimated probability that collisions are preventable by a skilled and attentive human driver strongly depends on the chosen human driver model.
As a proof of concept, the case study has used a simple well-known driver behavior model with few adaptations.
On the one hand, using a simple model contributes to the explainability of the results, ensures short simulation run times, and facilitates the reproducibility of the case study. 
On the other hand, the fidelity of the simulation results may be compromised by the simplicity of the human driver behavior model.
When using the proposed method to assess the risk of deploying \iac{ads} in the real world, evidence is needed to justify the fidelity of the simulation results.
More research is needed to actually verify the fidelity of the simulation results.

Whereas the case study has employed a human driver behavior model, it is also possible to perform simulations with a real human, i.e., with driver-in-the-loop simulations.
Given the number of simulation runs that has been conducted, it is impracticable to conduct all simulations with real humans, but a combination of driver-in-the-loop simulations and simulations with driver behavior models is possible. 
For example, the driver behavior models may be tuned to match the results of the driver-in-the-loop simulations such that the simulations with the driver behavior models can be further used to interpolate the driver-in-the-loop simulation results.

In addition to driver-in-the-loop simulations, the results of the extent to which collisions are reasonably preventable could be compared to accident data.
\cstart However, to truly compare the results of accident data with the results of the simulations of the skilled and attentive driver, only the accident data in which a skilled and attentive driver collided should be considered. 
Given a real-world accident, it might be challenging to determine, in hindsight, whether the drivers involved in the accident were skilled and attentive. 
The drivers' skills may be based on whether the drivers were (non-)professionals with certain number of years of experience, whether the drivers were involved in accidents before, etc.
Especially the attentiveness might be challenging to determine in hindsight, especially if the accidents were fatal.
Therefore, more research is needed to determine how and to which extent accident data can be used as a comparison of (driver-in-the-loop) simulations of skilled and attentive drivers. \cend

\acresetall
\section{Conclusions}
\label{sec:conclusions}

The proposal for a new United Nations regulation concerning the approval of \iac{ads} has been an important milestone toward the deployment of highly-automated vehicles.
This proposal has stated that ``the activated system shall not cause any collisions that are reasonably foreseeable and preventable.''
In this work, we have proposed novel methods to provide more clarity on the meaning of ``reasonably foreseeable and preventable''.
More specifically, the proposed methods provide two data-driven approaches to determine the scenarios that are ``reasonably foreseeable'' and, therefore, are to be considered during the development of \iac{ads}.
Furthermore, the proposed methods include two approaches to determine to which extent a skilled and attentive human driver is able to avoid collisions, i.e., to quantify whether or not in potentially critical scenarios, a collision is ``reasonably preventable''. 
The result can be used as a benchmark for \iac{ads}.

By means of a case study, the proposed methods have been illustrated.
The case study has considered three types of scenarios: scenarios with a leading vehicle decelerating, cut-in scenarios, and scenarios in which the ego vehicle approaches a slower leading vehicle. 
For each of these three scenario categories, we have determined the ranges of the parameter values for the reasonably foreseeable scenarios. 
Furthermore, the likelihood that a skilled and attentive human driver cannot avoid a collision has been estimated using simulations.

Future work involves applying the proposed method with more data, thereby providing more accurate results.
Additionally, it would be of interest to investigate appropriate values for the threshold that is used to determine the set of reasonably foreseeable scenarios.
Other future work involves investigating whether the chosen human driver behavior model is appropriate and, if needed, improving the human driver behavior models to better mimic the behavior of a skilled and attentive human driver.

\section*{Acknowledgment}
The work presented in this paper is part of the HEADSTART project. 
This project has received funding from the European Union's Horizon 2020 research and innovation programme under grant agreement No.\ 824309. 
Content reflects only the authors’ view and European Commission is not responsible for any use that may be made of the information it contains.

\bibliographystyle{abbrvnat}
\bibliography{../bib}

\begin{thebibliography}{43}
\providecommand{\natexlab}[1]{#1}
\providecommand{\url}[1]{\texttt{#1}}
\expandafter\ifx\csname urlstyle\endcsname\relax
  \providecommand{\doi}[1]{doi: #1}\else
  \providecommand{\doi}{doi: \begingroup \urlstyle{rm}\Url}\fi

\bibitem[{Aurora}(2019)]{aurora2019newera}
{Aurora}.
\newblock The new era of mobility.
\newblock Technical report, Aurora, 2019.
\newblock URL
  \url{https://downloads.ctfassets.net/v3j0gnq3qxwi/4QVMTwpBo2ZOmE03B09UP2/611de2c139aef05d7204ace06e946e00/VSSA_Final.pdf}.

\bibitem[Balkema and De~Haan(1974)]{balkema1974residual}
A.~A. Balkema and L.~De~Haan.
\newblock Residual life time at great age.
\newblock \emph{The Annals of Probability}, 2\penalty0 (5):\penalty0 792--804,
  1974.
\newblock \doi{10.1214/aop/1176996548}.

\bibitem[Bimbraw(2015)]{bimbraw2015autonomous}
K.~Bimbraw.
\newblock Autonomous cars: Past, present and future a review of the
  developments in the last century, the present scenario and the expected
  future of autonomous vehicle technology.
\newblock In \emph{12th International Conference on Informatics in Control,
  Automation and Robotics (ICINCO)}, volume~1, pages 191--198, 7 2015.
\newblock URL \url{https://ieeexplore.ieee.org/abstract/document/7350466}.

\bibitem[Chan(2017)]{chan2017advancements}
C.-Y. Chan.
\newblock Advancements, prospects, and impacts of automated driving systems.
\newblock \emph{International Journal of Transportation Science and
  Technology}, 6\penalty0 (3):\penalty0 208--216, 2017.
\newblock \doi{10.1016/j.ijtst.2017.07.008}.

\bibitem[de~Gelder and Paardekooper(2017)]{deGelder2017assessment}
E.~de~Gelder and J.-P. Paardekooper.
\newblock Assessment of automated driving systems using real-life scenarios.
\newblock In \emph{IEEE Intelligent Vehicles Symposium (IV)}, pages 589--594,
  2017.
\newblock \doi{10.1109/ivs.2017.7995782}.

\bibitem[de~Gelder et~al.(2020{\natexlab{a}})de~Gelder, Manders, Grappiolo,
  Paardekooper, Op~den Camp, and De~Schutter]{degelder2020scenariomining}
E.~de~Gelder, J.~Manders, C.~Grappiolo, J.-P. Paardekooper, O.~Op~den Camp, and
  B.~De~Schutter.
\newblock Real-world scenario mining for the assessment of automated vehicles.
\newblock In \emph{IEEE International Transportation Systems Conference
  (ITSC)}, pages 1073--1080, 2020{\natexlab{a}}.
\newblock \doi{10.1109/ITSC45102.2020.9294652}.

\bibitem[de~Gelder et~al.(2020{\natexlab{b}})de~Gelder, Op~den Camp, and
  de~Boer]{degelder2019scenariocategories}
E.~de~Gelder, O.~Op~den Camp, and N.~de~Boer.
\newblock Scenario categories for the assessment of automated vehicles.
\newblock Technical report, CETRAN, 2020{\natexlab{b}}.
\newblock URL
  \url{http://cetran.sg/wp-content/uploads/2020/01/REP200121_Scenario_Categories_v1.7.pdf}.
\newblock Version 1.7.

\bibitem[de~Gelder et~al.(2021)de~Gelder, Elrofai, Khabbaz~Saberi, Op~den Camp,
  Paardekooper, and De~Schutter]{degelder2021risk}
E.~de~Gelder, H.~Elrofai, A.~Khabbaz~Saberi, O.~Op~den Camp, J.-P.
  Paardekooper, and B.~De~Schutter.
\newblock Risk quantification for automated driving systems in real-world
  driving scenarios.
\newblock \emph{IEEE Access}, 9:\penalty0 168953--168970, 2021.
\newblock \doi{10.1109/ACCESS.2021.3136585}.

\bibitem[de~Gelder et~al.(2022)de~Gelder, Paardekooper, Khabbaz~Saberi,
  Elrofai, Op~den Camp, Kraines, Ploeg, and De~Schutter]{degelder2020ontology}
E.~de~Gelder, J.-P. Paardekooper, A.~Khabbaz~Saberi, H.~Elrofai, O.~Op~den
  Camp, S.~Kraines, J.~Ploeg, and B.~De~Schutter.
\newblock Towards an ontology for scenario definition for the assessment of
  automated vehicles: An object-oriented framework.
\newblock \emph{IEEE Transactions on Intelligent Vehicles}, 7\penalty0
  (2):\penalty0 300--314, 2022.
\newblock \doi{10.1109/TIV.2022.3144803}.

\bibitem[de~Haan(1994)]{dehaan1994extreme}
L.~de~Haan.
\newblock Extreme value statistics.
\newblock In \emph{Extreme Value Theory and Applications}, pages 93--122.
  Springer, 1994.

\bibitem[Duong(2007)]{duong2007ks}
T.~Duong.
\newblock {ks}: Kernel density estimation and kernel discriminant analysis for
  multivariate data in {R}.
\newblock \emph{Journal of Statistical Software}, 21\penalty0 (7):\penalty0
  1--16, 2007.
\newblock \doi{10.18637/jss.v021.i07}.

\bibitem[Duong(2015)]{duong2015spherically}
T.~Duong.
\newblock Spherically symmetric multivariate beta family kernels.
\newblock \emph{Statistics \& Probability Letters}, 104:\penalty0 141--145,
  2015.
\newblock \doi{10.1016/j.spl.2015.05.012}.

\bibitem[{ECE/TRANS/WP.29/2022/59/Rev.1}(2022)]{ece2022WP29}
{ECE/TRANS/WP.29/2022/59/Rev.1}.
\newblock Proposal for the 01 series of amendments to un regulation no.\ 157
  (automated lane keeping systems).
\newblock Standard, World Forum for Harmonization of Vehicle Regulations, 2022.
\newblock URL
  \url{https://unece.org/sites/default/files/2022-05/ECE-TRANS-WP.29-2022-59r1e.pdf}.

\bibitem[{E/ECE/TRANS/505/Rev.3/Add.156}(2021)]{ece2021WP29}
{E/ECE/TRANS/505/Rev.3/Add.156}.
\newblock Uniform provisions concerning the approval of vehicles with regard to
  automated lane keeping systems.
\newblock Standard, World Forum for Harmonization of Vehicle Regulations, 2021.
\newblock URL \url{https://unece.org/sites/default/files/2021-03/R157e.pdf}.

\bibitem[Elfring et~al.(2016)Elfring, Appeldoorn, van~den Dries, and
  Kwakkernaat]{elfring2016effective}
J.~Elfring, R.~Appeldoorn, S.~van~den Dries, and M.~Kwakkernaat.
\newblock Effective world modeling: Multisensor data fusion methodology for
  automated driving.
\newblock \emph{Sensors}, 16\penalty0 (10):\penalty0 1--27, 2016.
\newblock \doi{10.3390/s16101668}.

\bibitem[Franke et~al.(2004)Franke, H{\"a}rdle, and
  Hafner]{franke2004statistics}
J.~Franke, W.~K. H{\"a}rdle, and C.~M. Hafner.
\newblock \emph{Statistics of Financial Markets}, volume~2.
\newblock Springer, 2004.
\newblock \doi{10.1007/978-3-030-13751-9}.

\bibitem[{General Motors}(2018)]{gm2018selfdriving}
{General Motors}.
\newblock Self-driving safety report.
\newblock Technical report, General Motors, 2018.
\newblock URL
  \url{https://www.gm.com/content/dam/company/docs/us/en/gmcom/gmsafetyreport.pdf}.

\bibitem[Green(2000)]{green2000long}
M.~Green.
\newblock ``{H}ow long does it take to stop?'' {M}ethodological analysis of
  driver perception-brake times.
\newblock \emph{Transportation Human Factors}, 2\penalty0 (3):\penalty0
  195--216, 2000.
\newblock \doi{10.1207/sthf0203_1}.

\bibitem[Hayward(1972)]{hayward1972near}
J.~C. Hayward.
\newblock Near miss determination through use of a scale of danger.
\newblock Technical Report TTSC-7115, Pennsylvania State University, 1972.
\newblock URL
  \url{https://onlinepubs.trb.org/Onlinepubs/hrr/1972/384/384-004.pdf}.

\bibitem[Kusano et~al.(2022)Kusano, Beatty, Schnelle, Favaro, Crary, and
  Victor]{kusano2022collision}
K.~D. Kusano, K.~Beatty, S.~Schnelle, F.~Favaro, C.~Crary, and T.~Victor.
\newblock Collision avoidance testing of the waymo automated driving system.
\newblock \emph{arXiv preprint arXiv:2212.08148}, 2022.
\newblock URL \url{https://arxiv.org/abs/2212.08148}.

\bibitem[Madni(2018)]{madni2018autonomous}
A.~M. Madni.
\newblock Autonomous system-of-systems.
\newblock In \emph{Transdisciplinary Systems Engineering}, pages 161--186.
  Springer, 2018.
\newblock \doi{10.1007/978-3-319-62184-5_10}.

\bibitem[Mahdinia et~al.(2020)Mahdinia, Arvin, Khattak, and
  Ghiasi]{mahdinia2020safety}
I.~Mahdinia, R.~Arvin, A.~J. Khattak, and A.~Ghiasi.
\newblock Safety, energy, and emissions impacts of adaptive cruise control and
  cooperative adaptive cruise control.
\newblock \emph{Transportation Research Record}, 2674\penalty0 (6):\penalty0
  253--267, 2020.
\newblock \doi{10.1177/0361198120918572}.

\bibitem[Mammeri et~al.(2015)Mammeri, Lu, and Boukerche]{mammeri2015design}
A.~Mammeri, G.~Lu, and A.~Boukerche.
\newblock Design of lane keeping assist system for autonomous vehicles.
\newblock In \emph{7th International Conference on New Technologies, Mobility
  and Security (NTMS)}, pages 1--5, 2015.
\newblock \doi{10.1109/NTMS.2015.7266483}.

\bibitem[Mattas et~al.(2022)Mattas, Albano, Don{\`a}, Galassi, Suarez-Bertoa,
  Vass, and Ciuffo]{mattas2022driver}
K.~Mattas, G.~Albano, R.~Don{\`a}, M.~C. Galassi, R.~Suarez-Bertoa, S.~Vass,
  and B.~Ciuffo.
\newblock Driver models for the definition of safety requirements of automated
  vehicles in international regulations. application to motorway driving
  conditions.
\newblock \emph{Accident Analysis \& Prevention}, 174:\penalty0 106743, 2022.
\newblock \doi{10.1016/j.aap.2022.106743}.

\bibitem[Milakis et~al.(2016)Milakis, Snelder, van Arem, van Wee, and
  de~Almeida~Correia]{milakis2016scenarios}
D.~Milakis, M.~Snelder, B.~van Arem, B.~van Wee, and G.~H. de~Almeida~Correia.
\newblock Scenarios about development and implications of automated vehicles in
  the {N}etherlands.
\newblock In \emph{95th Annual Meeting Transportation Research Board}, 2016.
\newblock URL
  \url{https://www.researchgate.net/profile/Dimitris-Milakis/publication/288828248_Scenarios_about_development_and_implications_of_automated_vehicles_in_the_Netherlands/links/5684dad708ae1975839451ff/Scenarios-about-development-and-implications-of-automated-vehicles-in-the-Netherlands.pdf}.

\bibitem[Muslim et~al.(2023)Muslim, Endo, Imanaga, Kitajima, Uchida, Kitahara,
  Ozawa, Sato, and Nakamura]{muslim2023cut}
H.~Muslim, S.~Endo, H.~Imanaga, S.~Kitajima, N.~Uchida, E.~Kitahara, K.~Ozawa,
  H.~Sato, and H.~Nakamura.
\newblock Cut-out scenario generation with reasonability foreseeable parameter
  range from real highway dataset for autonomous vehicle assessment.
\newblock \emph{IEEE Access}, 11:\penalty0 45349--45363, 2023.
\newblock \doi{10.1109/ACCESS.2023.3268703}.

\bibitem[Najm et~al.(2007)Najm, Smith, and
  Yanagisawa]{USDoT2007precrashscenarios}
W.~G. Najm, J.~D. Smith, and M.~Yanagisawa.
\newblock Pre-crash scenario typology for crash avoidance research.
\newblock Technical Report DOT HS 810 767, U.S. Department of Transportation
  Research and Innovative Technology Administration, 4 2007.
\newblock URL \url{https://rosap.ntl.bts.gov/view/dot/6281/dot_6281_DS1.pdf}.

\bibitem[Nakamura et~al.(2022)Nakamura, Muslim, Kato, Pr{\'e}fontaine-Watanabe,
  Nakamura, Kaneko, Imanaga, Antona-Makoshi, Kitajima, Uchida, Kitahara, Ozawa,
  and Taniguchi]{nakamura2022defining}
H.~Nakamura, H.~Muslim, R.~Kato, S.~Pr{\'e}fontaine-Watanabe, H.~Nakamura,
  H.~Kaneko, H.~Imanaga, J.~Antona-Makoshi, S.~Kitajima, N.~Uchida,
  E.~Kitahara, K.~Ozawa, and S.~Taniguchi.
\newblock Defining reasonably foreseeable parameter ranges using real-world
  traffic data for scenario-based safety assessment of automated vehicles.
\newblock \emph{IEEE Access}, 10:\penalty0 37743--37760, 2022.
\newblock \doi{10.1109/ACCESS.2022.3162601}.

\bibitem[Owen(2013)]{owen2013montecarlo}
A.~B. Owen.
\newblock {M}onte {C}arlo theory, methods and examples, 2013.
\newblock URL \url{https://statweb.stanford.edu/~owen/mc/}.

\bibitem[Paardekooper et~al.(2019)Paardekooper, Montfort, Manders, Goos,
  de~Gelder, Op~den Camp, Bracquemond, and
  Thiolon]{paardekooper2019dataset6000km}
J.-P. Paardekooper, S.~Montfort, J.~Manders, J.~Goos, E.~de~Gelder, O.~Op~den
  Camp, A.~Bracquemond, and G.~Thiolon.
\newblock Automatic identification of critical scenarios in a public dataset of
  6000 km of public-road driving.
\newblock In \emph{26th International Technical Conference on the Enhanced
  Safety of Vehicles (ESV)}, 2019.
\newblock URL
  \url{https://www-esv.nhtsa.dot.gov/Proceedings/26/26ESV-000255.pdf}.

\bibitem[Parzen(1962)]{parzen1962estimation}
E.~Parzen.
\newblock On estimation of a probability density function and mode.
\newblock \emph{The Annals of Mathematical Statistics}, 33\penalty0
  (3):\penalty0 1065--1076, 1962.
\newblock \doi{10.1214/aoms/1177704472}.

\bibitem[Pickands(1975)]{pickands1975statistical}
J.~Pickands.
\newblock Statistical inference using extreme order statistics.
\newblock \emph{The Annals of Statistics}, pages 119--131, 1975.
\newblock \doi{10.1214/aos/1176343003}.

\bibitem[Rosenblatt(1956)]{rosenblatt1956remarks}
M.~Rosenblatt.
\newblock Remarks on some nonparametric estimates of a density function.
\newblock \emph{The Annals of Mathematical Statistics}, 27\penalty0
  (3):\penalty0 832--837, 1956.
\newblock \doi{10.1214/aoms/1177728190}.

\bibitem[{SAE J3016}(2021)]{sae2021j3016}
{SAE J3016}.
\newblock Taxonomy and definitions for terms related to driving automation
  systems for on-road motor vehicles.
\newblock Technical report, SAE International, 4 2021.

\bibitem[Schakel et~al.(2010)Schakel, van Arem, and Netten]{schakel2010effects}
W.~J. Schakel, B.~van Arem, and B.~D. Netten.
\newblock Effects of cooperative adaptive cruise control on traffic flow
  stability.
\newblock In \emph{13th International IEEE Conference on Intelligent
  Transportation Systems}, pages 759--764, 2010.
\newblock \doi{10.1109/itsc.2010.5625133}.

\bibitem[Schoener(2020)]{schoener2020challenging}
H.-P. Schoener.
\newblock Challenging highway scenarios beyond collision avoidance for
  autonomous vehicle certification.
\newblock \emph{Researchgate preprint}, 2020.
\newblock \doi{10.13140/RG.2.2.29355.05926}.

\bibitem[Silverman(1986)]{silverman1986density}
B.~W. Silverman.
\newblock \emph{Density Estimation for Statistics and Data Analysis}.
\newblock CRC press, 1986.

\bibitem[Tarko(2012)]{tarko2012use}
A.~P. Tarko.
\newblock Use of crash surrogates and exceedance statistics to estimate road
  safety.
\newblock \emph{Accident Analysis \& Prevention}, 45:\penalty0 230--240, 2012.
\newblock \doi{10.1016/j.aap.2011.07.008}.

\bibitem[Treiber et~al.(2000)Treiber, Hennecke, and
  Helbing]{treiber2000congested}
M.~Treiber, A.~Hennecke, and D.~Helbing.
\newblock Congested traffic states in empirical observations and microscopic
  simulations.
\newblock \emph{Physical review E}, 62\penalty0 (2):\penalty0 1805--1824, 2000.
\newblock \doi{10.1103/PhysRevE.62.1805}.

\bibitem[Turlach(1993)]{turlach1993bandwidthselection}
B.~A. Turlach.
\newblock Bandwidth selection in kernel density estimation: A review.
\newblock Technical report, Institut f{\"u}r Statistik und {\"O}konometrie,
  Humboldt-Universit{\"a}t zu Berlin, 1993.
\newblock URL
  \url{https://www.researchgate.net/publication/2316108_Bandwidth_Selection_in_Kernel_Density_Estimation_A_Review}.

\bibitem[Vellinga(2019)]{vellinga2019automated}
N.~E. Vellinga.
\newblock Automated driving and the future of traffic law.
\newblock In \emph{Regulating New Technologies in Uncertain Times}, pages
  67--82. Springer, 2019.
\newblock \doi{10.1007/978-94-6265-279-8_5}.

\bibitem[Waymo(2021)]{waymo2021safety}
Waymo.
\newblock Waymo safety report.
\newblock Technical report, Waymo, 2021.
\newblock URL
  \url{https://downloads.ctfassets.net/sv23gofxcuiz/4gZ7ZUxd4SRj1D1W6z3rpR/2ea16814cdb42f9e8eb34cae4f30b35d/2021-03-waymo-safety-report.pdf}.

\bibitem[Zhang(1996)]{zhang1996nonparametric}
P.~Zhang.
\newblock Nonparametric importance sampling.
\newblock \emph{Journal of the American Statistical Association}, 91\penalty0
  (435):\penalty0 1245--1253, 1996.
\newblock \doi{10.1080/01621459.1996.10476994}.

\end{thebibliography}

\end{document}